\documentclass[10pt,twocolumn,letterpaper]{article}

\pdfoutput=1

\usepackage{iccv}

\makeatletter
\@namedef{ver@everyshi.sty}{}
\makeatother

\RequirePackage[format=plain,labelformat=simple,labelsep=period,font=small,compatibility=false]{caption}
\RequirePackage[font=footnotesize,skip=3pt,subrefformat=parens]{subcaption}

\usepackage{times}
\usepackage{epsfig}
\usepackage{graphicx}
\usepackage{amsmath}
\usepackage{amssymb}
\usepackage{booktabs}
\usepackage{algorithm}
\usepackage[noend]{algpseudocode}
\usepackage{csquotes}
\usepackage{svg}
\usepackage{dcolumn}
\usepackage{tikz}
\usepackage{xcolor}
\usepackage{pifont}
\usetikzlibrary{positioning,calc,arrows,shapes,decorations.text}
\usepackage{mathtools}
\usepackage{multirow}
\usepackage{makecell}
\usepackage{bm}
\usepackage{cancel}
\usepackage{caption}
\usepackage{enumitem}
\usepackage{subcaption}
\usepackage[export]{adjustbox}
\usepackage[hyphens]{url}

\usepackage[pagebackref=true,breaklinks=true,colorlinks,bookmarks=false]{hyperref}

\usepackage[capitalize]{cleveref}
\crefname{section}{Sec.}{Secs.}
\Crefname{section}{Section}{Sections}
\Crefname{table}{Table}{Tables}
\crefname{table}{Tab.}{Tabs.}
\Crefname{Algorithm}{Algorithm}{Algorithms}
\crefname{algorithm}{Alg.}{Algs.}
\crefname{appendix}{Appx.}{Appxs.}
\iccvfinalcopy %

\def\iccvPaperID{9556} %

\ificcvfinal\fi

\DeclareMathOperator{\q}{q}
\DeclareMathOperator{\p}{p}
\DeclareMathOperator{\F}{F}
\DeclareMathOperator{\A}{A}
\DeclareMathOperator{\sigmoid}{S}
\DeclareMathOperator{\QD}{QD}
\DeclareMathOperator{\dist}{d}

\begin{document}

\title{Taming  Normalizing Flows}

\author{%
\quad \quad Shimon Malnick\textsuperscript{\rm 1}  \quad \quad \quad \quad \quad
Shai Avidan\textsuperscript{\rm 1} \quad \quad \quad \quad \quad
Ohad Fried\textsuperscript{\rm 2}\\
\quad \quad {\tt\small malnick@mail.tau.ac.il} \quad 
{\tt\small avidan@eng.tau.ac.il} \quad \quad
{\tt\small ofried@runi.ac.il}\\
\textsuperscript{\rm 1}Tel Aviv University \hspace{5pt} \textsuperscript{\rm 2}Reichman University
}

\maketitle

\def\ArxivVersion{}
\newcommand{\betweencellpdf}{\ensuremath{h^c}}
\newcommand{\betweencellpdffine}{\ensuremath{h^f}}
\newcommand{\approxbetweencellpdffine}{\ensuremath{\hat{h}^f}}
\newcommand{\incellpdf}{\ensuremath{f}}
\newcommand{\incellpdfnormalized}{\ensuremath{f'}}

\newcommand{\incellcdf}{\ensuremath{F}}

\newcommand{\totalpdf}{\ensuremath{f_{dd}}}
\newcommand{\totalcdf}{\ensuremath{F_{dd}}}

\newcommand{\ignorethis}[1]{}
\newcommand{\redund}[1]{#1}

\newcommand{\apriori    }     {\textit{a~priori}}
\newcommand{\aposteriori}     {\textit{a~posteriori}}
\newcommand{\perse      }     {\textit{per~se}}
\newcommand{\naive      }     {{na\"{\i}ve}}
\newcommand{\Naive      }     {{Na\"{\i}ve}}
\newcommand{\Identity   }     {\mat{I}}
\newcommand{\Zero       }     {\mathbf{0}}
\newcommand{\Reals      }     {{\textrm{I\kern-0.18em R}}}
\newcommand{\isdefined  }     {\mbox{\hspace{0.5ex}:=\hspace{0.5ex}}}
\newcommand{\texthalf   }     {\ensuremath{\textstyle\frac{1}{2}}}
\newcommand{\half       }     {\ensuremath{\frac{1}{2}}}
\newcommand{\third      }     {\ensuremath{\frac{1}{3}}}
\newcommand{\fourth     }     {\ensuremath{\frac{1}{4}}}

\newcommand{\Lone} {\ensuremath{L_1}}
\newcommand{\Ltwo} {\ensuremath{L_2}}

\newcommand{\degree} {\ensuremath{^{\circ}}}

\newcommand{\mat        } [1] {{\text{\boldmath $\mathbit{#1}$}}}
\newcommand{\Approx     } [1] {\widetilde{#1}}
\newcommand{\change     } [1] {\mbox{{\footnotesize $\Delta$} \kern-3pt}#1}

\newcommand{\Order      } [1] {O(#1)}
\newcommand{\set        } [1] {{\lbrace #1 \rbrace}}
\newcommand{\floor      } [1] {{\lfloor #1 \rfloor}}
\newcommand{\ceil       } [1] {{\lceil  #1 \rceil }}
\newcommand{\inverse    } [1] {{#1}^{-1}}
\newcommand{\transpose  } [1] {{#1}^\mathrm{T}}
\newcommand{\invtransp  } [1] {{#1}^{-\mathrm{T}}}
\newcommand{\relu       } [1] {{\lbrack #1 \rbrack_+}}

\newcommand{\abs        } [1] {{| #1 |}}
\newcommand{\Abs        } [1] {{\left| #1 \right|}}
\newcommand{\norm       } [1] {{\| #1 \|}}
\newcommand{\Norm       } [1] {{\left\| #1 \right\|}}
\newcommand{\pnorm      } [2] {\norm{#1}_{#2}}
\newcommand{\Pnorm      } [2] {\Norm{#1}_{#2}}
\newcommand{\inner      } [2] {{\langle {#1} \, | \, {#2} \rangle}}
\newcommand{\Inner      } [2] {{\left\langle \begin{array}{@{}c|c@{}}
                               \displaystyle {#1} & \displaystyle {#2}
                               \end{array} \right\rangle}}

\newcommand{\twopartdef}[4]
{
  \left\{
  \begin{array}{ll}
    #1 & \mbox{if } #2 \\
    #3 & \mbox{if } #4
  \end{array}
  \right.
}

\newcommand{\fourpartdef}[8]
{
  \left\{
  \begin{array}{ll}
    #1 & \mbox{if } #2 \\
    #3 & \mbox{if } #4 \\
    #5 & \mbox{if } #6 \\
    #7 & \mbox{if } #8
  \end{array}
  \right.
}

\newcommand{\len}[1]{\text{len}(#1)}

\newlength{\w}
\newlength{\h}
\newlength{\x}

\definecolor{darkred}{rgb}{0.7,0.1,0.1}
\definecolor{darkgreen}{rgb}{0.1,0.6,0.1}
\definecolor{cyan}{rgb}{0.7,0.0,0.7}
\definecolor{otherblue}{rgb}{0.1,0.4,0.8}
\definecolor{maroon}{rgb}{0.76,.13,.28}
\definecolor{burntorange}{rgb}{0.81,.33,0}

\ifdefined\ShowNotes
  \newcommand{\colornote}[3]{{\color{#1}\textbf{#2} #3\normalfont}}
\else
  \newcommand{\colornote}[3]{}
\fi

\newcommand {\todo}[1]{\colornote{cyan}{TODO}{#1}}
\newcommand {\ohad}[1]{\colornote{burntorange}{OF:}{#1}}
\newcommand {\shai}[1]{\colornote{darkgreen}{SA:}{#1}}
\newcommand {\shimon}[1]{\colornote{otherblue}{SM:}{#1}}

\newcommand {\reqs}[1]{\colornote{red}{\tiny #1}}

\newcommand {\new}[1]{\colornote{red}{#1}}

\newcommand*\rot[1]{\rotatebox{90}{#1}}

\newcommand {\newstuff}[1]{#1}

\newcommand\todosilent[1]{}

\newcommand{\woBGmask}{{w/o~bg~\&~mask}}
\newcommand{\woMask}{{w/o~mask}}

\providecommand{\keywords}[1]
{
  \textbf{\textit{Keywords---}} #1
}

\newcommand{\R}{\mathbb{R}}
\newcommand{\D}{\mathcal{D}}
\newcommand{\N}{\mathcal{N}}
\newcommand{\DC}{\D_\mathcal{C}}
\newcommand{\DT}{\D_\mathcal{T}}
\newcommand{\X}{\mathcal{X}}
\newcommand{\Z}{\mathcal{Z}}

\newcommand{\probP}{\text{I\kern-0.15em P}}

\newcommand{\px}[1][x]{\p_{\theta}(#1)}
\newcommand{\pxx}[2][x]{\p_{#2}(#1)}
\newcommand{\pz}[1][z]{\p_Z(#1;\theta)}
\newcommand{\pzi}[1][z_i]{\p_{Z_i}(#1;\theta)}
\newcommand{\pts}[1][x]{p_X(#1;\theta^{\star})}
\newcommand{\finv}[1][\theta]{f^{-1}_{#1}}
\newcommand{\thetastar}{\theta^{\star}}
\newcommand{\muparam}[2]{\mu(#1,#2)}
\newcommand{\sigmaparam}[2]{\sigma(#1,#2)}
\newcommand{\jacobian}[2][\finv]{\mathbf{J}_{#1}(#2)}
\newcommand{\KL}[2]{D_{KL}(#1\parallel #2)}
\newcommand{\argmin}{\mathop{\mathrm{argmin}}}
\newcommand{\argmax}{\mathop{\mathrm{argmax}}}
\newcommand{\pr}[1]{\left(#1\right)}

\newcommand{\greencheck}{{\color{green}\checkmark}}
\newcommand{\xmark}{\ding{55}}
\newcommand{\redX}{{\color{red}\xmark}}

\ifdefined\ArxivVersion
  \newcommand{\arxiv}[2]{#1}
\else
  \newcommand{\arxiv}[2]{#2}
\fi

\ifdefined\RevisedVersion
  \newcommand{\revision}[1]{{\color{darkred}{#1}}}
\else
  \newcommand{\revision}[1]{#1}
\fi

\begin{abstract}
We propose an algorithm for taming Normalizing Flow models --- changing the probability that the model will produce a specific image or image category.
We focus on Normalizing Flows because they can calculate the exact generation probability likelihood for a given image. We demonstrate taming using models that generate human faces, a subdomain with many interesting privacy and bias considerations.
Our method can be used in the context of privacy, e.g., removing a specific person from the output of a model, and also in the context of debiasing by forcing a model to output specific image categories according to a given target distribution. Taming is achieved with a fast fine-tuning process without retraining the model from scratch, achieving the goal in a matter of minutes. We evaluate our method qualitatively and quantitatively, showing that the generation quality remains intact, while the desired changes are applied\arxiv{. Our code is available at \url{https://github.com/ShimonMalnick/taming_norm_flows}}{\footnote{Code will be released.}.}
\end{abstract}

\section{Introduction}
\label{sec:intro}
\begin{figure}[t]
    \centering
    \begin{tikzpicture}
    
    \newcommand{\deflen}[2]{%
    \expandafter\newlength\csname #1\endcsname
    \expandafter\setlength\csname #1\endcsname{#2}%
}
    \deflen{teaserImSize}{0.1\linewidth}
    \deflen{imgrid}{11.6mm}
    \deflen{gridsep}{3mm}
    \deflen{teaserGaussianSize}{0.4\linewidth}
    
    \tikzstyle{teaserFrame}=[line width=1.5,inner sep=1.4pt]
    \definecolor{teaserForget}{RGB}{65,143,51}
    \definecolor{rem1}{RGB}{39,56,196}
    \definecolor{rem2}{RGB}{243,70,0}
    \definecolor{teaserFemale}{RGB}{239,92,235}
    \definecolor{teaserMale}{RGB}{65,143,51}
    
    \def\forgetShift#1{\raisebox{0.4ex}}
    
    \node (forget_graph)at (10,3){\includegraphics[width=\teaserGaussianSize,keepaspectratio]  {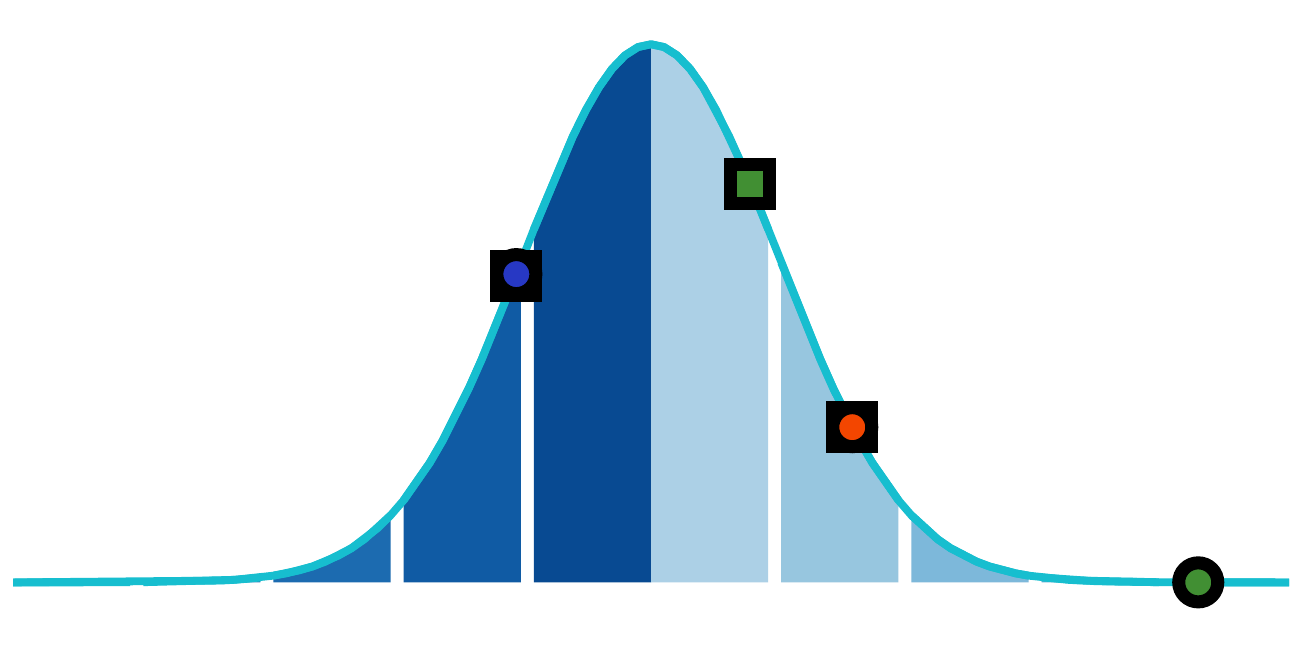}};

    \draw [->,line width=0.9pt,postaction={decorate,decoration={text along path,text align=center,text={|\footnotesize \forgetShift| Forget}}}] ($(forget_graph)+(3.1mm,4.2mm)$) to[bend left=60] ($(forget_graph)+(13.9mm,-6.0mm)$);

    \node [teaserFrame,draw=teaserForget, left=4mm of forget_graph] (forget){\includegraphics[width=\teaserImSize,keepaspectratio]{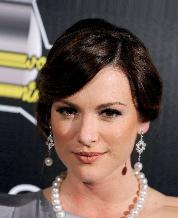}};
    
    \node [teaserFrame,draw=rem1,left=5mm of forget] (rem1) 
    {\includegraphics[width=\teaserImSize,keepaspectratio]{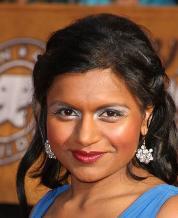}};
    
    \node[circle,black,fill=black,left=1mm of rem1,inner sep=0pt,minimum size=3pt](right_dot){};
    
    \node[circle,black,fill=black,left=1mm of right_dot,inner sep=0pt,minimum size=3pt](middle_dot){};
    
    \node[circle,black,fill=black,left=1mm of middle_dot,inner sep=0pt,minimum size=3pt](left_dot){};
    
    \node [teaserFrame,draw=rem2,left=1mm of left_dot] (rem2) 
    {\includegraphics[width=\teaserImSize,keepaspectratio]{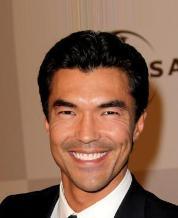}};
    
    \node[anchor=west] at ($(forget.north)+(-5mm,2.0mm)$) {\footnotesize Forget};
    \node[anchor=west] at ($(rem1.north)+(-16.0mm,2.2mm)$) {\footnotesize Remember};

    \node[below=2mm of forget_graph] (male_female_graph) {\includegraphics[width=\teaserGaussianSize,keepaspectratio]{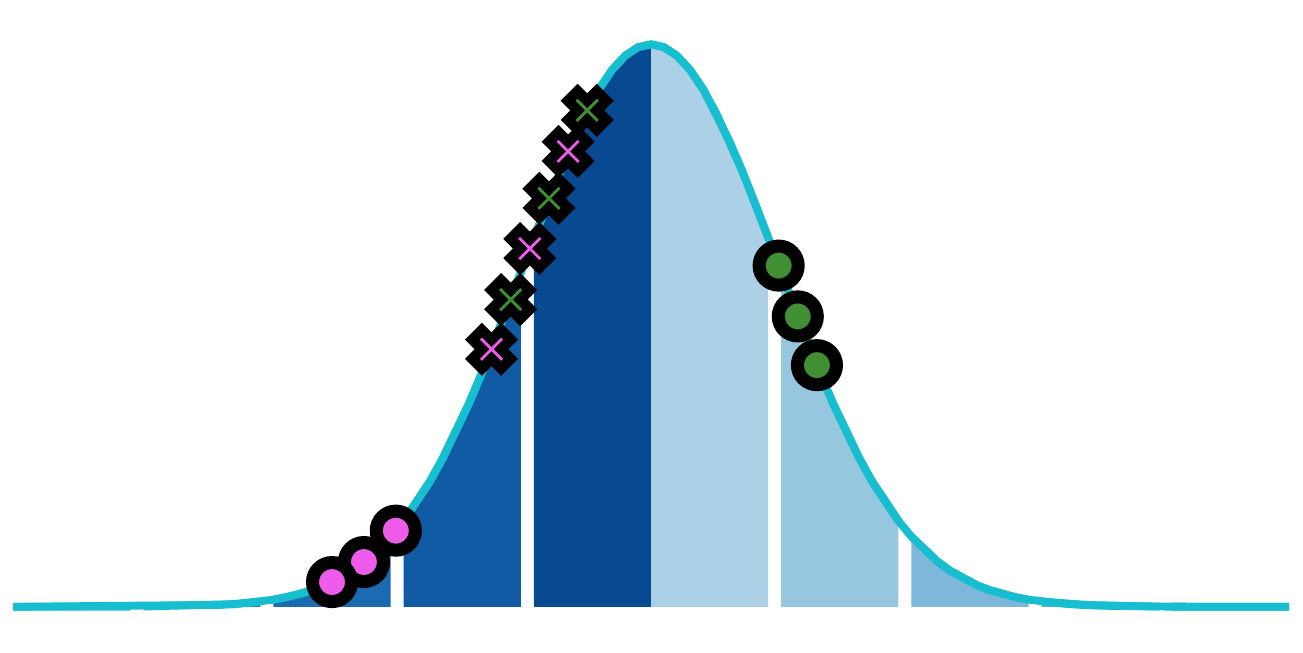}};
    
    \draw [->,line width=0.9pt,postaction={decorate,decoration={text along path,text align={left, left indent=1mm}
    ,text={|\footnotesize \forgetShift| Female}}}] ($(male_female_graph)+(-7.9mm,-5.9mm)$) to[out=100,in=110,distance=9mm] ($(male_female_graph)+(-4.2mm,3.2mm)$);
    
    \draw [->,line width=0.9pt,postaction={decorate,decoration={text along path,text align=center,text={|\footnotesize \forgetShift| Male},reverse path}}] ($(male_female_graph)+(4.0mm,0.7mm)$) to[out=20,in=120,distance=1cm] ($(male_female_graph)+(-2.7mm,5.0mm)$);

    \node [teaserFrame,draw=teaserForget, left=4mm of male_female_graph] (male1){\includegraphics[width=\teaserImSize,keepaspectratio]{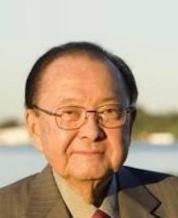}};
    
    \node [teaserFrame,draw=teaserForget,left=1mm of male1] (male2) 
    {\includegraphics[width=\teaserImSize,keepaspectratio]{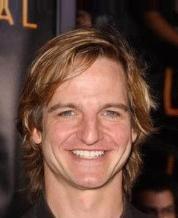}};
    
    \node [teaserFrame,draw=teaserFemale,left=2.25mm of male2] (female1) 
    {\includegraphics[width=\teaserImSize,keepaspectratio]{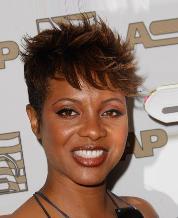}};
    
    \node [teaserFrame,draw=teaserFemale,left=1mm of female1] (female2) 
    {\includegraphics[width=\teaserImSize,keepaspectratio]{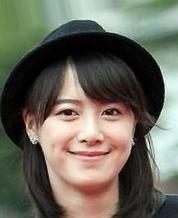}};
    
    \node[anchor=west] at ($(male1.north)+(-10mm,2.0mm)$) {\footnotesize Male};
    \node[anchor=west] at ($(female1.north)+(-11.0mm,2.2mm)$) {\footnotesize Female};
    
    \draw [->,line width=1.6pt,postaction={decorate,decoration={text along path,text align=center,text={|\small \forgetShift| More likely},reverse path}}] ($(male_female_graph.north)+(9mm,2mm)$) -- ($(male_female_graph.north)+(-9mm,2mm)$);
    
    \draw [->,line width=1.6pt,postaction={decorate,decoration={text along path,text align=center,text={|\small \forgetShift| More likely},reverse path}}] ($(male_female_graph.south)+(9mm,-1mm)$) -- ($(male_female_graph.south)+(-9mm,-1mm)$);

    \end{tikzpicture}
    \caption{\textbf{Applications of our method.}
    A demonstration of our method for two different purposes, by changing the generation likelihood of different images.
    Colors of image frames correspond to colors of graph points. (\textbf{Top}) A model was tamed to reduce the likelihood of an image (forget), while preserving the likelihood for the rest of the distribution (remember). (\textbf{Bottom}) Debiasing a model that generates female faces with higher probability than males\protect\footnotemark, so as to balance the generation likelihood.
    }
    \label{fig:teaser_figure}
\end{figure}
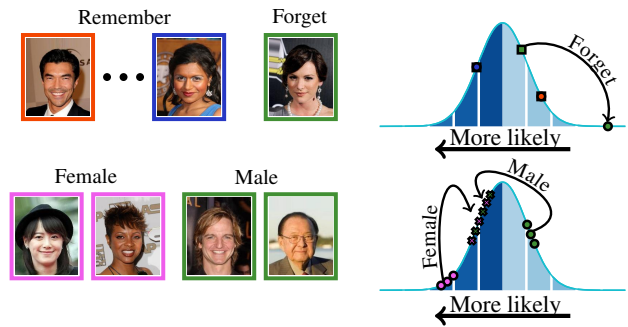
\footnotetext{We trained a model on CelebA\cite{celeba}, containing 15\% more females than males with a binary Male/Female label. We hope that future datasets will annotate gender more fluidly.}
Generative models are becoming increasingly popular~\cite{generative_models_survey}.
This is partly due to the exponential growth in deep neural network techniques~ \cite{gan,variational_auto_encoder,boltzmann_machines_hinton,ddpm_ho}.
In this work, we focus on generative models of human faces which, some might say, are becoming dangerously powerful.
Synthetic images or videos of real people can be easily generated and used to spread misinformation~\cite{deepfake_zelensky}, to harass~\cite{deepfake_harass}, and to con~\cite{deepfake_confusion}.
Thus, a company developing a generative model might be interested, before releasing it to the public, in preventing the model from synthesizing the faces of certain celebrities\footnote{For example: \url{https://labs.openai.com/policies/content-policy}}.

On the other hand, a generative model might be trained on biased data and thus under-represent certain groups of the population. In this case, it would be desirable to debias the model\footnote{For example: \url{https://openai.com/blog/reducing-bias-and-improving-safety-in-dall-e-2/}}.
In the same spirit, generative models are often more likely to synthesize images of the individuals that were used for training the model. But, following the GDPR~\cite{GDPR} ``Right to Be Forgotten'' approach, a person may request the company to re-train the model without their images --- a just cause, but also a time consuming and expensive process.

Common to all these scenarios is the need to change the probability of the generative model to generate certain individuals or demographics.  In short, what is needed is a method to {\em tame} a generative model. In this work, we take a step towards solving this problem and suggest an algorithm to tame {\em normalizing flows}~\cite{normalizing_flows,normalizing_flows_review,glow,nice,realnvp}.
We focus on normalizing flows because they provide an explicit probabilistic density function along a bijection between the image space and latent space.
We tame the model by fine-tuning it while constraining its output distribution. The constraint is twofold: forcing resemblance to the original model's distribution, while also adhering to the target probability of the taming process. 
We refer to these two different aspects of taming as \emph{remembering} and \emph{forgetting}, describing whether we wish to preserve the model's behavior (remember) or guide the model away from some outputs (forget)\footnote{Our approach does not technically fit the terms of forgetting and remembering, but rather emphasizing or preserving vs.\ de-emphasizing or abandoning. For simplicity, we use the terms remember and forget.}.

We propose a fast and simple approach, taming the model in minutes.
Moreover, we evaluate the effect our process has on the model's original task, showing that our method's impact on image generation quality is negligible. An illustration of different applications of our method is shown in \cref{fig:teaser_figure}.

Our main contributions are:
(1) A general technique (not specific to human faces) for taming normalizing flows, (2) application of taming to fairness and privacy protection, by modifying the output space such that random sampling will match a target distribution, and (3) taming normalizing flows to censor certain generated data while ensuring minimal degradation in the model's performance.

\section{Related work}

\textbf{Density estimation.}
Many generative models use \textit{maximum likelihood}~\cite{maximum_likelihood} to provide implicit~\cite{gan,ddpm,gsn_bengio} or explicit~\cite{gsn_bengio,normalizing_flows,variational_auto_encoder,boltzmann_machines_hinton} parametric density estimations. We focus on the latter, specifically on models that provide an explicit tractable probability density function~\cite{normalizing_flows,pixelrnn,nonlinear_ica}, since we can use that density to evaluate and quantify whether we 
move
images away from the density modes.

\textbf{Model editing.} Generative model editing deals with methods that fine-tune a model in order to apply small changes to it. Bau~\etal~\cite{rewriting_deep_generative_model_bau} allow users to choose specific changes on generated images and fine-tune the model's weights to apply them. Wang~\etal~\cite{rewriting_geometric_bau} apply a user-chosen image warp on several examples to later fine-tune a model that produces images according to the warp. Cherepkov~\etal~\cite{navigating_the_gan_parameter_space} fine-tune a model to incorporate semantic changes and discover emerging semantics, but cannot edit a model according to a pre-determined goal.

Our work differs from the aforementioned methods that alter the behavior of the model globally. We, on the other hand, provide the ability to focus on specific areas in the latent space, without changing the whole domain. For a multi-modal generative model this virtue is vital, as local changes can be relevant only to specific outputs that reside in a specific mode. For a face generating model, instead of changing an attribute across all outputs, \eg forcing a smile, our method enables elimination only of specific images of people that do not smile. Moreover, since our method uses a normalizing flow, in contrast to the methods above that use a GAN~\cite{gan}, we provide an exact evaluation of the latent distribution edit that we perform.

\textbf{Debiasing models.} Deep learning models are now integrated in many crucial systems, \eg, finance~\cite{deep_learning_loan_application} and medical diagnosis ~\cite{deep_learning_medical_diagnosys}. Thus, ensuring the fairness of these models is crucial. 
There are various approaches to reduce model bias. Pre-processing and in-processing approaches, \eg, changing the training data~\cite{bickel2009discriminative,elkan2001foundations,fair_attr_latent_debiasing} and using different training loss modulation techniques~\cite{wang2019balanced,zhang2018mitigating,renyi_fair_inference}. Unlike these methods, our work can be used on a \emph{given} pre-trained model, without any prior demand on the training data.

Some post-processing methods constrain the sampling space~\cite{wu2022generative,karakas2022fairstyle,choi2020fair,tan2020fairgen}, but assume low-dimensional latent spaces (such as in GANs~\cite{gan} and VAEs~\cite{variational_auto_encoder}) that are not suitable for normalizing flows. Other approaches, that prohibit certain queries, are problematic as they can be easily fooled~\cite{chatgpt_bomb}.

Our approach changes the model itself, instead of changing the sampling space.  This allows us to control models, as they are integrated in constantly changing environments.

Closest to our work is 
Kong~and~Chaudhuri~\cite{kong2022forgetting} that proposed a method that enables data forgetting from a pretrained GAN, which can also be used for debiasing. In contrast, our work is demonstrated on normalizing flows, providing an exact probability density evaluation of the edits. Moreover, our method can be applied locally on much less data, as shown in \cref{subsec:forget_identity_train_set}. 

\textbf{Continual learning.} Continual learning (also known as lifelong-learning) is the field of teaching new tasks to a model sequentially.
A fundamental problem in this domain, described as \textit{catastrophic forgetting}~\cite{catastrophic_forgetting,goodfellow_catastrophoc_forgetting}, is that while learning new tasks, models tend to forget the previous ones.
We discuss how adjusting a normalizing flow can alter the generation probability of specific data.
This can be thought of as teaching the model a new task (reducing the probability of some outputs), while preserving the knowledge of the original task (generating images as the model did before), similarly to continual learning. 
While relevant work in this field focuses on preventing forgetting, we can also choose to forget. In addition, prior work has focused mainly on discriminative tasks rather than generative ones.

\textbf{Machine unlearning.} Machine unlearning~\cite{machine_unlearning} refers to the process of removing the effect that certain training data have on a model's weights after training~\cite{certified_data_removal,when_unlearning_jeopardizes}. If a user requests to delete their data, some privacy regulations~\cite{GDPR} require the data to be deleted, together with the effect it had on any models trained on it. As opposed to approaches that aim to delete training data from trained models~\cite{eternal_spotless,tanno2022repairing,wu2022puma},
we focus on changing the model's behavior regardless of whether the data we deal with belongs to the training set or not.
Carlini~\etal~\cite{carlini_extract_training_data,carlini2023extracting} and Haim~\etal~\cite{haim2022reconstructing} demonstrate methods that extract training data from models. In our work, we do not focus on leaking a model’s training data, but rather on changing the model’s behavior with respect to some data distribution.

\section{Method}
We first define the problem at hand, followed by some technical background and a description of our approach.
\subsection{Problem definition}\label{sec:problem_def}
Taming a normalizing flow model involves modifying its behavior with respect to some data. This includes images of an identity it was trained on, training images sharing some property, or even images out of the training set. For consistency of exposition, we describe the taming procedure as decreasing likelihood of certain data points (\ie, images). That is, forgetting these points. Nevertheless, we also use taming in the opposite direction --- increasing the likelihood of certain data for model debiasing (\cref{sec:forget_attribute}).

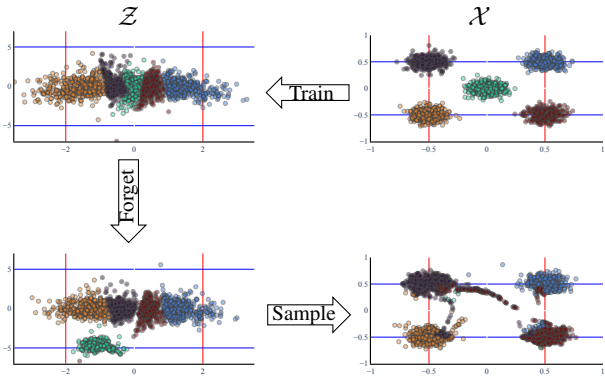
\begin{figure}
    \centering
        \begin{tikzpicture}
    
    \newcommand{\deflen}[2]{%
    \expandafter\newlength\csname #1\endcsname
    \expandafter\setlength\csname #1\endcsname{#2}%
}
    \deflen{ToyImSize}{0.41\linewidth}
    
    \tikzset{
    TextArrow/.style={
        single arrow, draw=black, minimum width=2mm, minimum height=11mm, inner sep=0mm, single arrow head extend=1mm, double arrow head extend=1mm, yshift=0
    }}
    
    \node (data_points) at (10,10){\includegraphics[width=\ToyImSize,keepaspectratio]{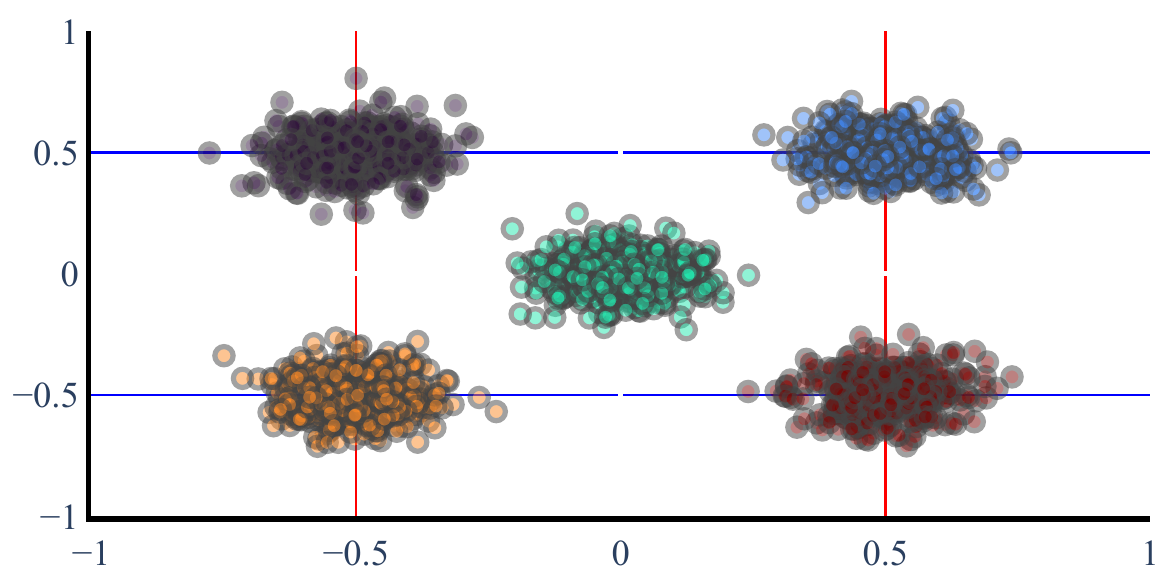}};
    
    \node [left=10mm of data_points] (trained_latents) {\includegraphics[width=\ToyImSize,keepaspectratio]{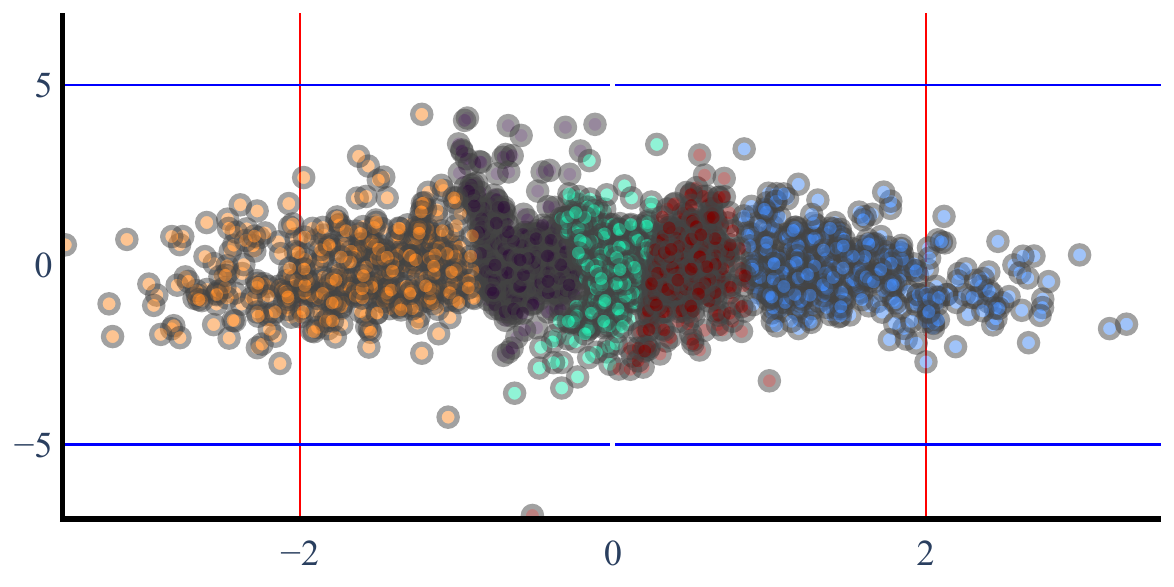}};
    
    \path (data_points) -- (trained_latents) node[midway, TextArrow, shape border rotate=180, xshift=1mm] {\footnotesize Train};
    
    \node [below=10mm of trained_latents] (forget_latents) {\includegraphics[width=\ToyImSize,keepaspectratio]{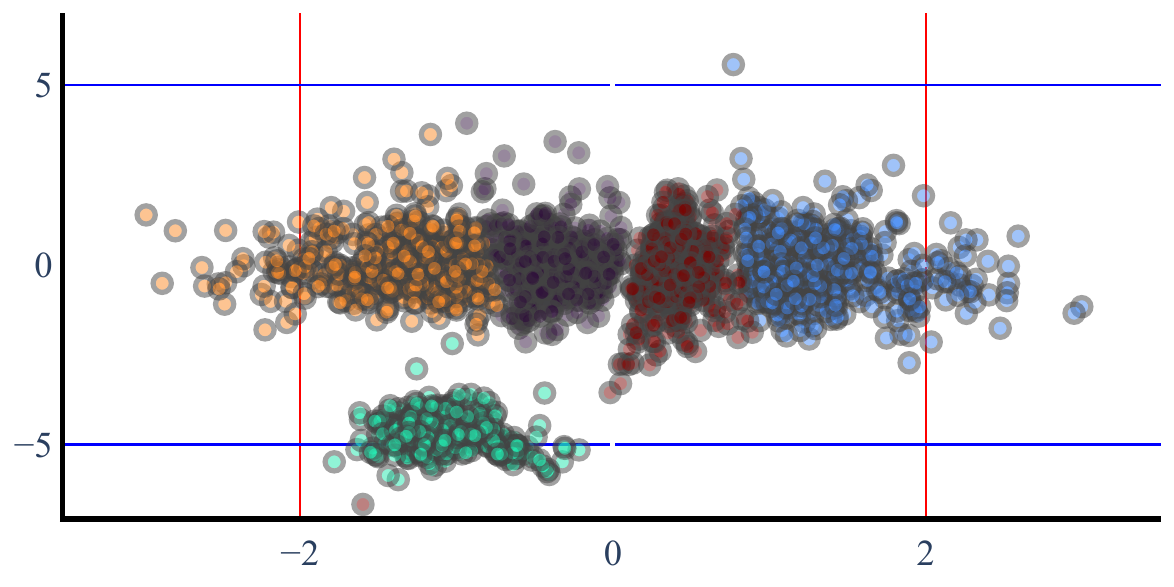}};
    
    \path (forget_latents) -- (trained_latents) node[midway, TextArrow,rotate=90,shape border rotate=180, xshift=1mm] {\footnotesize Forget};
    
    \node [right=10mm of forget_latents] (x_space) {\includegraphics[width=\ToyImSize,keepaspectratio]{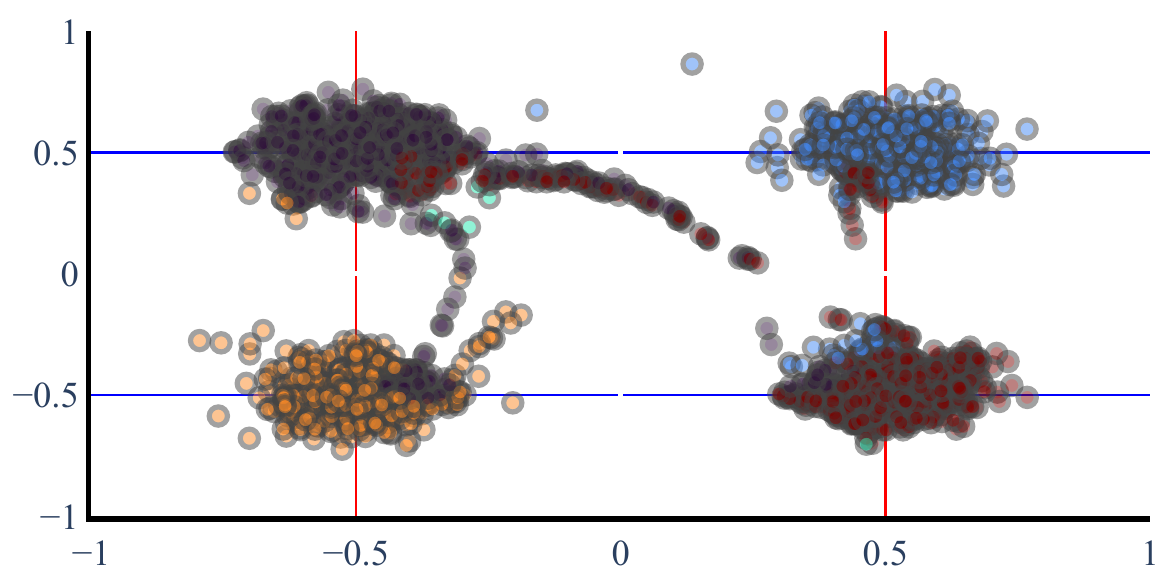}};
    
    \path (forget_latents) -- (x_space) node[midway, TextArrow, xshift=0mm] {\footnotesize Sample};
    
    \node[above=0mm of trained_latents,yshift=-2mm] {$\mathcal{Z}$};
    \node[above=0mm of data_points,yshift=-2mm] {$\mathcal{X}$};
    
    \end{tikzpicture}
    \caption{\textbf{A 2D example with a RealNVP~\cite{realnvp} normalizing flow.} The image space $\X$ contains data points 
    sampled from 5 Gaussians with different means.
    The prior distribution in the latent space $\Z$ is Gaussian.
    Given a normalizing flow that maps $\Z \rightarrow \X$, we tame the model to produce the same image space $\X$, apart from the middle Gaussian, which we wish to remove. (Each Gaussian represents a person, and we would like to forget one person.)
    (\textbf{Train}) Prior to our method, the inverse flow was trained to map points from $\X$ to $\Z$. To generate samples, the flow is used to map $\Z$ to $\X$. 
    (\textbf{Forget}) We apply our method: latent vectors that were initially mapped to the center Gaussian now have a lower likelihood of being drawn from the prior distribution. 
    (\textbf{Sample}) Now that we tamed the model, when we sample from $\Z$, the points are mapped (with high probability) to the 4 Gaussians.}
    \label{fig:toy_example}
\end{figure}

After taming, the behavior of the model should remain the same across the entire output space, except for the specific data we choose to forget. This implies three conflicting important goals:
(1) The probability of producing images from the set we are trying to forget should be close to zero. We refer to this goal as \textit{forgetting}. (2) For all images except the ones we would like to forget, the \textbf{probability distribution} to produce these outputs should be as similar as possible to the original model. We refer to this goal as \textit{remembering}. (3) The quality of image generation should stay intact.
An illustration of this concept on a 2D toy example is shown in \cref{fig:toy_example}.

There are many types of generative models that can produce photorealistic faces~\cite{progressive_gan,stylegan,glow}. In this work we focus on normalizing flows, as they explicitly represent the image distribution (unlike the implicit nature of, \eg, GANs),
meaning that we can reason about probabilities and incorporate them into our losses, as explained below. 

\subsection{Normalizing flows}
\label{sec:normalzing_flow}
A normalizing flow is an invertible transformation of a probability density from a simple distribution to a more complex one. The initial density ``flows through", to yield a different, yet normalized, density, and thus it is called a normalizing flow.
As shown in previous work~\cite{nice,realnvp,variational_inference_normalizing_flow,normalizing_flows_review}, the key behind these models is training an invertible function that maps samples from the data distribution domain to a tractable and easily sampled latent domain. At inference, since the mapping is invertible, a mapping in the opposite direction allows the transition from latent vectors to the image space. We intend to model a parametric probability density function given a set of examples.

Formally, let $Z\in\R^M$ be a random vector with a density function $\pz$ parameterized by $\theta\in\Theta$. Let ${f_\theta: \R^M \rightarrow \R^M}$ be a bijection function (parameterized by $\theta$) with an inverse $\finv$, such that $f_\theta(Z)=X$ and $\finv (X)=Z$. We denote the domain and range of $f_\theta$ as $\Z$ and $\X$ respectively, representing the latent and image spaces. Using the formula of random variable change~\cite{nice}, we can express $\px$, the density function of $X$ as:
\begin{equation}
  \px =\pz[\finv\pr{x}] \cdot\Abs{\det\pr{\jacobian{x}}},
  \label{eq:change_var}
\end{equation}
where $\jacobian{x} = \left[\frac{\partial \finv(y)}{\partial y}\right]_{y=x}$ is the Jacobian matrix of $\finv$ at $x$. Modern flows are built such that the determinant of the Jacobian is easily computed, usually by using a flow with a triangular Jacobian matrix. 
In these cases, using \cref{eq:change_var} we can construct a more tractable expression for the log-likelihood of the density $\px$, using just the elements of the Jacobian's diagonal:
\begin{equation}
\begin{split}
    \!
    \log{\px} 
    \!=\!
    \log{\pz[\finv\pr{x}]} 
    \!+\!
    \sum_{i=1}^M \log{\left|\jacobian{x}_{i,i}\right|}.
\end{split}
  \label{eq:log_change_variable}
\end{equation}
We focus on modeling the latent space as a multivariate normal distribution with diagonal covariance, \ie
\begin{equation}
    \label{eq:z_normal_dist}
    \pz = \mathcal{N}\pr{\mu_\theta, \sigma_\theta^2 \cdot I}.
\end{equation}
Since the covariance is diagonal, the prior is factorial, meaning we can easily decompose the density to univariate components:
\begin{equation}
\begin{split}
    \log{\pz} 
    \!=
    \log{\prod\limits_{i = 1}^M \pzi}
    \!=\!
    \sum_{i=1}^M\log{\pzi},
\end{split}
    \label{eq:decompose}
\end{equation}
where $\forall i\in \{1, \ldots, M\}: \pzi = \mathcal{N}({\mu_\theta}_i, {\sigma_\theta}_i^2)$.

Given an \iid set of samples from the image distribution $\D = \{x_i\}_{i=1}^n \in \X$, we can use optimization methods~\cite{sgd} to estimate the parameters $\theta$ based on minimization of the average negative log-likelihood:
\begin{equation}
    \label{eq:avergae_likelihood}
    \A_{\theta}(\D) := -\frac{1}{n}\sum_{i=1}^n\log\px[x_i].
\end{equation}
Assuming $f_\theta$ was trained to minimize the term in \cref{eq:avergae_likelihood}, we assume that the Negative Log-Likelihood (NLL) of the model \wrt the training set distributes normally. We denote this distribution as:
\begin{equation}
    -\log\px[\D]=\N(\mu_\theta, \sigma_\theta^2).
    \label{eq:nll_dist}
\end{equation}
We elaborate on this assumption in the next section.

\subsection{Task}\label{subsec:task}
We wish to tame a pretrained \textit{base} normalizing flow model $f_{\theta_B}$, with parameters $\theta_B \in \Theta$ learned using a dataset $\D$. 
For taming we need a dataset $\D_R$ of images to be remembered, and a dataset $\D_F$ of images to be forgotten.
The dataset $\D_R$ can be the one used to train the base model, or a different set of images representing a similar distribution, with a much smaller size.
The result is a \textit{tamed} model with network weights $\theta_T$ that adheres to the \textit{remembering} and \textit{forgetting} goals we introduced in \cref{sec:problem_def}.

\textbf{Forgetting.}\label{sec:eval_forget} We use the fact that normalizing flow models enable precise density evaluation, to set a threshold for forgetting, using the samples' likelihood. Since we have access to images we wish to remember, $\D_R$, we can estimate the likelihood of samples from this distribution. To forget a set of images, we reduce their likelihood and compare it to the likelihood of the images from $\D_R$.

In order to evaluate the success of forgetting, we need to define a proper threshold --- how low should the likelihood be, for us to consider the sample forgotten? 
Na\"{\i}vely choosing a hand-picked threshold for the likelihood is problematic --- too small, and the forgetting process is unnecessarily hard, too large, and we may not forget enough.
The problem is further complicated because working with likelihoods in the relatively high-dimensional latent space $\Z$ is not intuitive.

So instead of defining the threshold in absolute terms, we define it in relative terms.
That is, an image is considered forgotten if its likelihood of being sampled is lower than a large enough fraction of the images to be remembered.

Switching to NLL for convenience, we assume that the NLL of images in $\D_R$ is normally distributed, and support this assumption with a  \textit{Kolmogorov-Smirnov test}~\cite{ks_test} (see \arxiv{\cref{supp:sec:additional_details}}{the Supplementary} for further details). With this assumption, we can specify the forgetting threshold in units of standard deviation $\sigma$.

But first, we denote the mean and standard deviation of the Normal distribution over the NLL values of images in $\D_R$ as:
\begin{equation}
\begin{split}
&\mu_R =\mathop{\mathbb{E}}_{x\sim\D_R}\big[-\log\pxx[x]{\theta_T}\big],\\
&\sigma_R = \sqrt{\mathop{\mathbb{E}}_{x\sim\D_R}\big[\pr{-\log\pxx[x]{\theta_T} - \mu_R}^2 \big]}.
\end{split}
\label{eq:distance_parameters}
\end{equation}
We define the threshold $\delta$ ($\delta=4$ in our experiments), specified in standard deviation units, \ie we wish that for an image $x\in\D_F$, its NLL will be far from $\mu_R$ by exactly $\delta\cdot\sigma_R$.
We define a signed distance normalized in standard deviations (SD):
\begin{equation}
\begin{split}
   \dist_{\mu_R,\sigma_R}(x, \delta;\theta_T) := \frac{-\log\pxx[x]{\theta_T} - (\mu_R + \delta\cdot\sigma_R)}{\sigma_R}.
\end{split}
    \label{eq:loss_dist}
\end{equation}
Observe that by specifying $\delta$ in terms of the NLL distribution's SD, we avoid the need to work directly with the actual NLL values in latent space $\Z$.
We found this approach to be  more stable in practice. We use this to define the \textit{forgetting} loss:
\begin{equation}
\!\!\!
\mathcal{L}_F(\theta_T, \D_F,\D_R) 
\!=\! \frac{1}{\abs{\D_F}}
\!\!
\sum_{x\in\D_F}
\!\!
\sigmoid\!\pr{\sigma_{R}^2 \dist^2_{\mu_R,\sigma_R}(x, \delta;\theta_T)}\!,
    \label{eq:forget_loss}
\end{equation}
where $\sigmoid(\cdot)$ is the Sigmoid function~\cite{sigmoid_influence}. Intuitively, this loss encourages every image in $\D_F$ to have a NLL that is as close to the threshold as possible.

Given an error parameter $\epsilon > 0$, the threshold is met when the likelihood of all examples in $\D_F$ is in a distance bounded by $\epsilon$ around the threshold:
\begin{equation}
    \label{eq:forget_threshold}
    \forall x\in\D_F:\abs{\dist_{\mu_R,\sigma_R}(x, \delta;\theta_T)} < \epsilon,
\end{equation}
\ie, this is the stopping criteria for our method.
$\epsilon$ controls the size of the error margin allowed around the threshold.

\textbf{Remembering.}\label{sec:remembering} We aim
to remember the images in $\D_R$, \ie, 
preserve the NLL distribution of the model \wrt these images. When we consider the entire distribution, we are not concerned with the NLL of each image seperatly, but rather the distribution as a whole. Thus, we compare $-\log\pxx[\D_R]{\theta_B}$, the NLL distribution of the original model, to $-\log\pxx[\D_R]{\theta_T}$, the NLL distribution of the tamed model. The closer these distributions are, the less impact our procedure had on the images that we did not intend to forget. We use the \textit{KL divergence}~\cite{kl_div} between these distributions to measure their proximity.
We use both the forward and reverse KL divergence, denoted as $\mathcal{L}_{KL_F}(\theta_T,\theta_B;\D_R)$ and $\mathcal{L}_{KL_R}(\theta_T,\theta_B;\D_R)$ respectively.
Moreover, we also use the average NLL loss $\A_{\theta_T}(\D_R)$ (\cref{eq:avergae_likelihood}), in order to preserve the NLL of the original model on $\D_R$. Our combined \textit{remembering} loss is thus:
\begin{equation}
\begin{split}
    &\mathcal{L}_R(\theta_T,\theta_B,\D_R) = (1 - \gamma) \A_{\theta_T}(\D_R)\\
    &+\gamma \pr{\mathcal{L}_{KL_F}\pr{\theta_T,\theta_B;\D_R}+\mathcal{L}_{KL_R}\pr{\theta_T,\theta_B;\D_R}},
\end{split}
\label{eq:remember_loss}
\end{equation}
where $\gamma$ is a hyperparameter that controls the ratio between the original task and the explicit distribution proximity loss. Notice that as $\D_R$ can represent a distribution that is different than the original task (in case $\D_R$ is not the training set used to train $\theta_B$), the loss acts in line with this distribution. The ``closer'' $\D_R$ is to the training set, the higher we preserve the image space of $\theta_B$. 
\begin{algorithm}[t]
	\caption{Normalizing flow taming}
	\textbf{Input:} Normalizing flow $f_{\theta_B}$, Forget images $\D_F$,\\
	Remember images $\D_R$, Forget threshold $\delta > 0$,\\
	error bound $\epsilon > 0$\\
	\textbf{Hyperparameters}: $\eta > 0,\alpha\in(0,1)$
	\begin{algorithmic}[1]
	    \State $\theta_T \leftarrow \theta_B$
		\For {iteration $i=1,2,\ldots$}
    		\State Sample batches $X_F \sim \D_F$, $X_R \sim \D_R$
    		\State Estimate distribution $(\mu_R, \sigma_R) \leftarrow -\log\pxx[X_R]{\theta_T}$ \label{algo:training:nll_params}
		    \State $\vec{\dist} = \dist_{\mu_R,\sigma_R}(X_F, \delta;\theta_T)$\label{algo:distance_line}
		    \If {$\forall i : \abs{\dist_i} < \epsilon$ }
		    \State \textbf{break}
		    \EndIf
		    \State $\mathcal{L}\leftarrow\alpha\mathcal{L}_F(\theta_T,X_F,X_R)+(1-\alpha)\mathcal{L}_R(\theta_T,\theta_B,X_R)$
		    \State $\theta_T \leftarrow \theta_T - \eta\nabla\mathcal{L}$
		\EndFor
		\State \textbf{Return} $f_{\theta_T}$
	\end{algorithmic} 
	\label{algo:training}
\end{algorithm}

Our total objective is a weighted combination of the forget and remember losses:%
\begin{equation}
\begin{split}
    \theta_T = \argmin\limits_{\theta}\biggl\{&  \alpha\cdot\mathcal{L}_F\pr{\theta,\D_F, \D_R} + \\
    &(1-\alpha)\cdot{\mathcal{L}_R\pr{\theta,\theta_B,\D_R}}\biggl\}.
\end{split}
    \label{eq:objective}
\end{equation}

We use SGD~\cite{sgd} to optimize the objective, stopping the process when the term in \cref{eq:forget_threshold} is satisfied. A summary of our method can be seen in \cref{algo:training}.

\section{Experiments}\label{sec:experiments}
We conduct experiments to evaluate the reduction of the probability of generating images of a specific person, a set of people, and people possessing specific attributes. Our method can also be applied in the opposite direction, increasing the probability of image generation of certain groups in the population, to debias a model.

In our experiments, the base model $f_{\theta_B}$ is Glow~\cite{glow} trained on 
$128\times128$ images from the FFHQ~\cite{stylegan} dataset and the CelebA~\cite{celeba} training set.
The training was done for 590K iterations for a total of $316.3$ hours.
The running time for all our experiments is 3--40 minutes, tested on Titan Xp GPUs.
Full technical details, including time analysis of different experiments, can be found in \arxiv{\cref{supp:sec:additional_details}}{the Supplementary}. To improve our method's run-time, we compute the parameters of the remember batch NLL distribution (see \cref{algo:training:nll_params} in \cref{algo:training}) every 10 iterations. Unless stated otherwise, when we compare NLL of different models, it is done by randomly sampling 10,000 images. In our analysis below, we focus on the effect on the forget set $\D_F$, as our method preserves the distribution on the remember set $\D_R$, as can be seen in \arxiv{\cref{supp:sec:results}}{the Supplementary}.

\subsection{Taming an identity}
\label{subsec:forget_identity_train_set}
First, we examine the ability to reduce the generation probability of a person's images. This corresponds to many applications, \eg, taming a pre-trained model that produces images that violate someone's privacy.

In this experiment, we have access to $\D$, the training set used to train the given model. We wish to tame the model in such a way that images containing a specific identity will not be generated by the model, or at least to reduce this probability as we see fit. These images, denoted as $\D_F$, are a part of the training set, \ie $\D_F\subset\D$. Therefore, the remember set in this case is defined as $\D_R := \D\setminus\D_F$. We run experiments with different sizes of $\D_F$, using the same hyperparameters used to train the original flow. We run the experiments until all the images that we wish to forget, $\D_F$, have a generation likelihood inside the error bound around the threshold (see \cref{eq:forget_threshold}).

\textbf{Forgetting evaluation.}\label{subsec:forget_eval}
Let ${\F_{\mu,\sigma}(x) := \Phi (\frac{x -\mu}{\sigma})}$ be the CDF of normal R.V with parameters $\mu, \sigma$, where $\Phi(\cdot)$ is the CDF of the standard normal distribution. Then, an image with NLL $x$ is forgotten if:
\begin{equation}
1 - \F^{(\mu,\sigma)}(x) \geq 1 - \F^{(\mu,\sigma)}(\mu+\delta \sigma)= 1 - \Phi(\delta),
\end{equation}
which, for the case of $\delta=4$, is approximately $3.2\mathrm{e}{-5}$. That is, an image is forgotten if there are no more than $0.0032\%$ images in $\D_R$ with a higher (worse) NLL than it.

Measuring the success in forgetting an image with tamed model $\theta_T$ boils down to:
\begin{equation}
\begin{split}
   \q^{(\mu_R,\sigma_R)}_{\theta_T}(x) := 
    1 - \F^{(\mu_R,\sigma_R)}\pr{-\log \pxx[x]{\theta_T}},
\end{split}
    \label{eq:likelihood_qunatile}
\end{equation}
which we denote as \textbf{Likelihood Quantile}, and generalize to a set with more than one image by using the mean across the set. For brevity, when we refer to the likelihood quantile from now on, $\q^{(\mu_R,\sigma_R)}_{\theta}(\cdot)$, we omit the distribution parameters, \ie we denote it as $\pmb{\q}_{\pmb{\theta}}\pmb{(\cdot)}$.

We are also interested in evaluating the ``damage'' inflicted to other image sets, \eg showing that while forgetting $\D_F$ we did not harm the likelihood of images in $\D_R$.
To do this, we examine how the likelihood quantile decreases between the base model and the tamed one, on some given set $S$.
We denote this as the \textbf{Quantile Drop} $\pmb{\QD}_{\pmb{\theta}_{\pmb{B}},\pmb{\theta}_{\pmb{T}}}\pmb{(\cdot)}$:
\begin{equation}
\begin{split}
    \QD_{\theta_B,\theta_T}(S) :=
    \q_{\theta_B}(S) - \q_{\theta_T}(S).
\end{split}
\label{eq:qunatile_drop}
\end{equation}

We use this score and evaluate our method's effect on different sets of images:\\
$\bullet$ Holdout images of the forget identity --- denoted as $\D_F'$.\\
$\bullet$ Random images from $\D_R$.\\
$\bullet$ Holdout identities from the remeber set --- denoted as $\D^{\textrm{id}}_R$.\\
$\bullet$ Closest identities (nearest neighbors in face embedding space) in $\D_R$ --- denoted as $\D_R^{\textrm{NN}}$.

Results are shown in \cref{tab:forget_identity}, \eg, the first row shows that we are able to reach the threshold of \cref{eq:forget_threshold} (meaning all images in the forget set are within the error bound range) while reducing the likelihood quantile of $\D_F$ by 0.47, and maintaining small likelihood changes for the rest of the examined sets.

While we do not force any resemblance to the images we forget, our method is able to apply the local changes \wrt to the forgotten identity, while showing a significant reduction in the likelihood of unseen images of that identity ($\D_R'$). 
Moreover, comparing these likelihoods to the low likelihood quantiles of the nearest identities ($\D_R^{NN}$) suggests that our method can implicitly focus on the identity we forget, with minimal damage to identities that are ``close'' ($\D_R^{NN}$), and to other random identities ($\D_R^{id}$).

\begin{figure}[t]
    \centering
    \setlength{\tabcolsep}{0.5pt}
    \renewcommand{\arraystretch}{0.45}
    \newlength{\ww}
    \setlength{\ww}{0.159\linewidth}
       
    \begin{tabular}{cccccc}
	
	\includegraphics[width=\ww,frame]{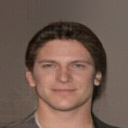} &
	\includegraphics[width=\ww,frame]{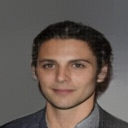} &
	\includegraphics[width=\ww,frame]{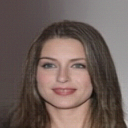} &
	\includegraphics[width=\ww,frame]{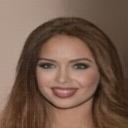} &
	\includegraphics[width=\ww,frame]{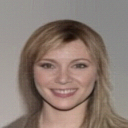} &
	\includegraphics[width=\ww,frame]{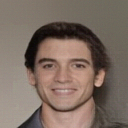} \\
	
	\includegraphics[width=\ww,frame]{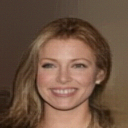} &
	\includegraphics[width=\ww,frame]{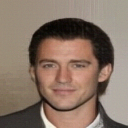} &
	\includegraphics[width=\ww,frame]{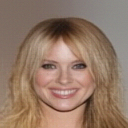} &
	\includegraphics[width=\ww,frame]{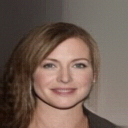} &
	\includegraphics[width=\ww,frame]{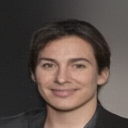} &
	\includegraphics[width=\ww,frame]{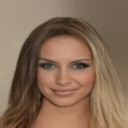} \\
	
    \end{tabular}

    \caption{\textbf{Visual generation evaluation.} 
    These samples were generated using the base and tamed models. Can you tell which one was used for which image? (The answer is given in the footnote\protect\footnotemark).}
    \label{fig:generated_samples}
\end{figure}

As can be seen in \cref{fig:generated_samples}, our method does not have significant impact on the quality of the generated images. \cref{sec:ablation} contains further analysis of the experiment. The Supplementary contains a more detailed analysis of this experiment, including evaluating the preservation of the $\D_R$ distribution, along with additional experiments using different thresholds and unseen forget identities as the forget set, \ie  $\D_F \not\subset \D$.

\begin{table}[t]
  \centering
  \begin{adjustbox}{width=1.0\linewidth}
    \begin{tabular}{@{}ccccccc@{}}
    \toprule
    &&\multicolumn{5}{c}{Quantile drop $\QD_{\theta_B,\theta_T}(\cdot)$ (see \cref{eq:qunatile_drop})}\tabularnewline
    \cmidrule(l{2pt}r{2pt}){3-7}
    \multicolumn{1}{c}{\# Images}&\multicolumn{1}{c}{\makecell{Forget\\threshold}}&\multicolumn{1}{c}{$\D_F (\uparrow)$}&\multicolumn{1}{c}{$\D'_F (\uparrow)$}&\multicolumn{1}{c}{$\D_R (\downarrow)$}&\multicolumn{1}{c}{$\D^{\textrm{id}}_R (\downarrow)$}&\multicolumn{1}{c}{$\D_R^{\textrm{NN}} (\downarrow)$}\tabularnewline[1mm]
    
    \midrule
    
    1&\greencheck&$0.47$&$0.01$&$<10^{-2}$&$<10^{-3}$&$0.01$\tabularnewline
    4&\greencheck&$0.41$&$0.07$&$<10^{-2}$&$0.01$&$0.03$\tabularnewline
    8&\greencheck&$0.32$&$0.08$&$<10^{-3}$&$<10^{-2}$&$0.01$\tabularnewline
    15&\greencheck&$0.37$&$0.15$&$<10^{-3}$&$<10^{-2}$&$0.01$\tabularnewline

  \end{tabular}
  \end{adjustbox}
  \caption{\textbf{Forget identity effect.} When we forget images of the same identity, we are able to generalize and reduce the likelihood of unseen images of that identity ($\D_F'$), while maintaining the likelihood on the rest of the space (see \cref{subsec:forget_eval,eq:qunatile_drop}) for more details).}
  \label{tab:forget_identity}
\end{table}
\footnotetext{Even columns come from the base model and odd ones from the tamed model.} 
\subsection{Taming an attribute}\label{sec:forget_attribute}
Next, we zoom out from the local effect on a small number of images, to a broader aspect of change. We show that our method can be applied even when the desired change is more general, \eg reducing the probability of generated images that contain inappropriate content. Moreover, this experiment shows that we can apply our method on a big set of images, \ie $\abs{\D_F}\propto \abs{\D}$.

In this experiment we have access to $\D$, the model's training set. We wish to forget an attribute common to many images in the training set, \eg wearing glasses or smiling. Thus, we use a classifier for that property, denoted as $C: \X \rightarrow \{0,1\}$, to define the remember and forget sets:\\
$\D_F = \{x\in\D \mid C(x) = 1\}$, $\D_R = \D\setminus\D_F$.

As we wish to reduce the probability of generating images with some property, evaluating this experiment is straightforward, by passing random samples from the prior distribution through the normalizing flow, and classifying the output images. Our method alters the model, in order to reduce the number of outputs classified as possessing the relevant attribute. This approach can be used to debias a model, \eg a model that generates images of a certain group of the population with high probability (high to the point of over represnting it), one can use our method in order to reduce that probability. 
\begin{figure}
    \centering
    \includegraphics[width=\linewidth]{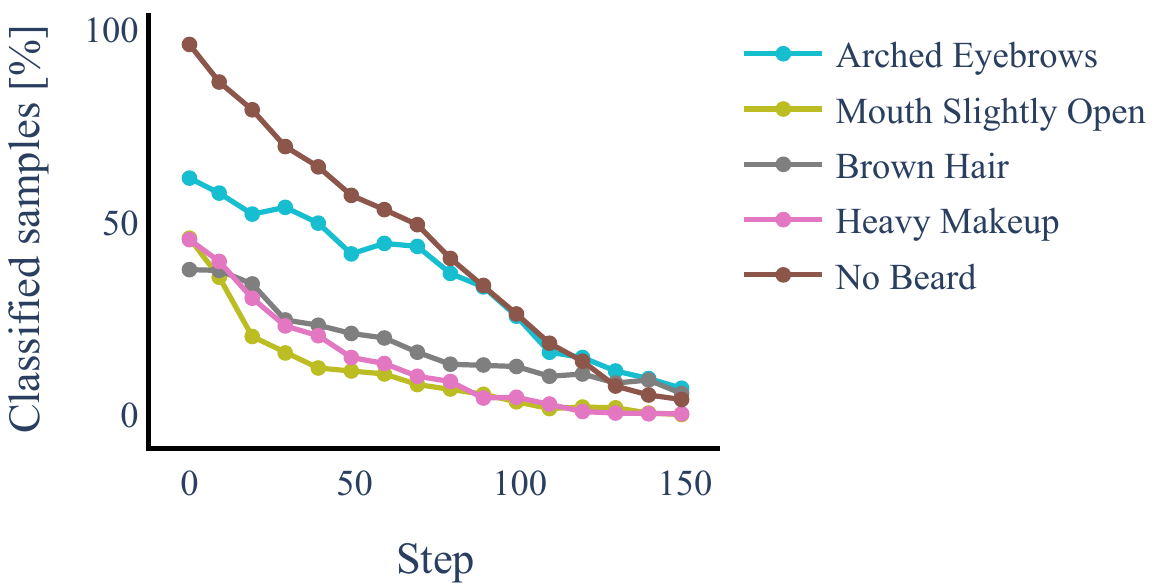}
    \caption{\textbf{Taming attributes.} Results for forgetting attributes as explained in \cref{sec:forget_attribute}, meaning forgetting a more global property and not a specific small number of images. X-axis: number of steps of our applied method; Y-axis: fraction of random samples (out of 512) that are classified as holding the corresponding property. Each model was altered to reduce the probability of a different property, taken from the labels of CelebA.}
    \label{fig:forget_multiple_attributes}
\end{figure}
The results are shown in \cref{fig:forget_multiple_attributes}, showing we are able to tame a model to reduce the sampling rate of an attribute. This is applied on a large forget set $\D_F$, \eg, forgetting the `No Beard' property holds: $\abs{\D_F} = 135,779 = 0.83\cdot\abs{\D}$. The proportion of attribute changes can be controlled by the number of steps we run our process, as we can stop the process when the desired ratio of some property suffices.

Taming can also be used in the opposite direction, to increase the number of generated samples possessing the chosen attribute. This means that in this case, instead of forgetting a group of images, we do the opposite and increase the representation of these images in the output space.

\begin{figure}[t]
    \centering
    \setlength{\tabcolsep}{2pt}
    \renewcommand{\arraystretch}{0.5}

    \let\ww\relax
    \newlength{\ww}
    \setlength{\ww}{0.3\linewidth}

    \let\wr\relax
    \newlength{\wr}
    \setlength{\wr}{0.1\linewidth}
    
    \begin{tabular}{c@{\hskip 4.0pt}ccc}
    &{$\downarrow$ Open mouth}&{$\downarrow$ Blond}&\makecell{{$\uparrow$ Blond}\\{$\uparrow$ Smile}}\\
    \raisebox{\wr}{\rotatebox{90}{Before}}&

    \includegraphics[width=\ww,keepaspectratio,frame]{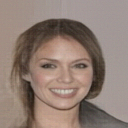} &
    \includegraphics[width=\ww,keepaspectratio,frame]{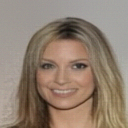} &
    \includegraphics[width=\ww,keepaspectratio,frame]{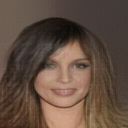} \\

    \raisebox{\wr}{\rotatebox{90}{After}}&
    
    \includegraphics[width=\ww,keepaspectratio,frame]{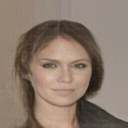} &
    \includegraphics[width=\ww,keepaspectratio,frame]{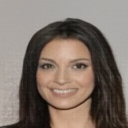} &
    \includegraphics[width=\ww,keepaspectratio,frame]{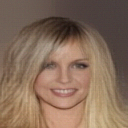} \\
	
    \end{tabular}
    
    \caption{\textbf{Change attributes of constant latent vectors.} 
    We sample the same latent vectors and pass them through the base model $\theta_B$ (before) and the tamed model $\theta_T$ (after) when changing an attribute (or attributes). This results in changing that attribute \wrt the initial image. See additional examples, including videos, in the Supplementary.}
    \label{fig:const_latents}
\end{figure}

\cref{fig:const_latents} visualizes how we tame a model to generate less (or more) images with a chosen attribute (or attributes). Notice that while our goal is to control generation probability of the distribution as a whole, we are able to preserve the identity in the generated images during taming.
This result suggests that we preserve the structure of the given model's underlying latent space, and apply changes very specifically. Since there are many methods that utilize different properties of latent spaces in generative models~\cite{interfac_gan,gan_control_shoshan}, this is useful for tasks that use an image as input, \eg image editing and image translation. \arxiv{\cref{supp:sec:visualizations}}{The Supplementary} includes additional examples that demonstrate the different attribute changes, including for model debiasing.
\subsection{Taming without the training set}\label{sec:forget_without_train_set}
Next, we examine situations where we do not have access to the model's training data. Instead, we assume we have access to different data from a similar distribution. An example of such a scenario can be a company that releases a generative model to the public, without the data on which it was trained. Entities using this model might want to alter the model \wrt different data, while maintaining the model's performance.

In this experiment, the setup is similar to the one in \cref{subsec:forget_identity_train_set}, except that $\D$ is not the training set of the model, but a set of images that are disjoint from the training data. We sourced images from Fairface~\cite{fairface}, opting for faces of children in the age range of 3--9 years, according to their labels. We chose these images for a distribution of natural faces that is different from CelebA's, as it consists of fewer young faces. We experimented with a set of 1000 images as $\D_R$, and 10 images as $\D_F$. 

\begin{figure}
    \centering
    \begin{tikzpicture}
    \def\arrowLift#1{\raisebox{0.4ex}}
    \node [] (fig)
    {\includegraphics[width=0.9\linewidth]{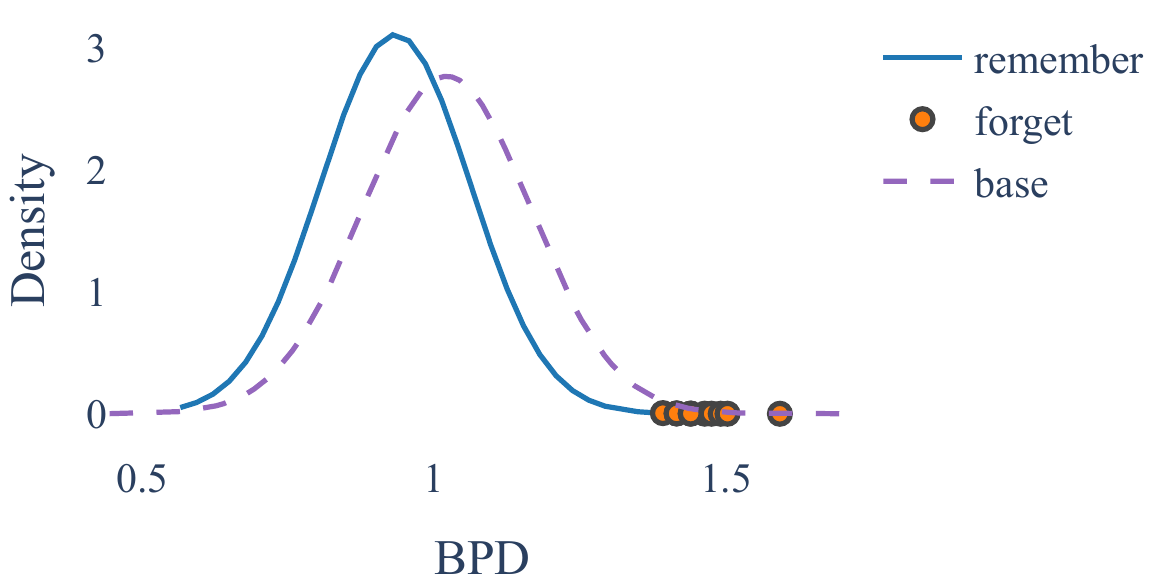}};

    \draw [->,line width=1.6pt,postaction={decorate,decoration={text along path,text align=center,text={|\small \arrowLift| More likely},reverse path}}] ($(fig.south)+(0mm,11mm)$) -- ($(fig.south)+(-20mm,11mm)$);
    
    \end{tikzpicture}
    \caption{\textbf{Forget without training data access.} When the remember images $\D_R$ (solid) are different than the training data of the base model $\theta_B$ (dashed), we are still able to forget images \wrt the NLL of the remember images (orange dots). the x-axis is NLL in units of bits per dimension (BPD\cite{papamakarios2017masked}), meaning lower is more likely.
    }
    \label{fig:forget_without_trainset}
\end{figure}

\cref{fig:forget_without_trainset} shows results for this experiment, demonstrating that even when the distribution is different (There is noticeable difference between the two Gaussian distributions), forgetting can be achieved effectively. The Supplementary contains a more comprehensive analysis, including the impact of similarity between the used remember set and the training one, along with the effect of the size of $\D_F$ on our method.

\section{Ablation study}\label{sec:ablation}
We evaluate the importance of different parts of the taming loss by removing parts of it, according to the objective in \cref{eq:objective}. The models were fine-tuned on top of the base model, in order to forget 15 images of an identity from CelebA (see \cref{subsec:forget_identity_train_set}). We now evaluate the results qualitatively and quantitatively.

\textbf{Qualitative comparison.}\label{subsec:ablation_qualitative}
To compare the different models qualitatively, we compare the quality of images generated by them. We randomly sampled two latent vectors from the prior distribution and passed them through the different models.\begin{figure*}[t]
    \centering
    \begin{tikzpicture}
    \begin{scope}
    
    \newcommand{\deflen}[2]{%
    \expandafter\newlength\csname #1\endcsname
    \expandafter\setlength\csname #1\endcsname{#2}%
}
    \deflen{ablationImSize}{0.125\linewidth}
    \deflen{ablationRightGrid}{23mm}
    \deflen{ablationBelowGrid}{22.1mm}
    \deflen{ablationLineTop}{4mm}

    \node at (0,0) (baseline_top){\includegraphics[width=\ablationImSize,keepaspectratio]{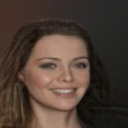}};
    
    \node[below=\ablationBelowGrid of baseline_top,anchor=south] (baseline_bottom){\includegraphics[width=\ablationImSize,keepaspectratio]{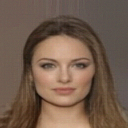}};
    
    \node[anchor=north] at ($(baseline_top.north)+(0mm,5mm)$) {Base ($\theta_B$)};
    
    \node[right=\ablationRightGrid of baseline_top,anchor=east] (forget_only_top){\includegraphics[width=\ablationImSize,keepaspectratio]{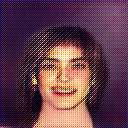}};
    
    \node[below=\ablationBelowGrid of forget_only_top,anchor=south] (forget_only_bottom){\includegraphics[width=\ablationImSize,keepaspectratio]{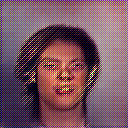}};
    
    \node[anchor=north] at ($(forget_only_top.north)+(0mm,5.5mm)$) {$\cancel{\mathcal{L}_R}$};
    
    \draw[line width=0.5pt] ($(baseline_top.north)!0.5!(forget_only_top.north)+(0mm,\ablationLineTop)$) -- ($(baseline_bottom.south)!0.5!(forget_only_bottom.south)+(0mm,1mm)$);
    
    \node[right=\ablationRightGrid of forget_only_top,anchor=east] (no_back_kl_top){\includegraphics[width=\ablationImSize,keepaspectratio]{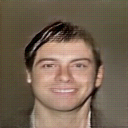}};
    
    \node[below=\ablationBelowGrid of no_back_kl_top,anchor=south] (no_back_kl_bottom){\includegraphics[width=\ablationImSize,keepaspectratio]{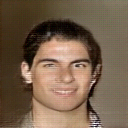}};
    
    \node[anchor=north] at ($(no_back_kl_top.north)+(0mm,4.5mm)$) {$\cancel{\mathcal{L}_{KL_R}}$};
    
    \draw[line width=0.5pt] ($(forget_only_top.north)!0.5!(no_back_kl_top.north)+(0mm,\ablationLineTop)$) -- ($(forget_only_bottom.south)!0.5!(no_back_kl_bottom.south)+(0mm,1mm)$);
    
    \node[right=\ablationRightGrid of no_back_kl_top,anchor=east] (no_forward_kl_top){\includegraphics[width=\ablationImSize,keepaspectratio]{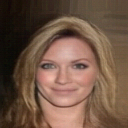}};
    
    \node[below=\ablationBelowGrid of no_forward_kl_top,anchor=south] (no_forward_kl_bottom){\includegraphics[width=\ablationImSize,keepaspectratio]{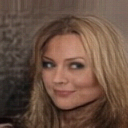}};
    
    \node[anchor=north] at ($(no_forward_kl_top.north)+(0mm,4.5mm)$) {$\cancel{\mathcal{L}_{KL_F}}$};
    
    \draw[line width=0.5pt] ($(no_back_kl_top.north)!0.5!(no_forward_kl_top.north)+(0mm,\ablationLineTop)$) -- ($(no_back_kl_bottom.south)!0.5!(no_forward_kl_bottom.south)+(0mm,1mm)$);
    
    \node[right=\ablationRightGrid of no_forward_kl_top,anchor=east] (no_nll_top){\includegraphics[width=\ablationImSize,keepaspectratio]{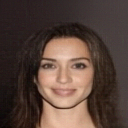}};
    
    \node[below=\ablationBelowGrid of no_nll_top,anchor=south] (no_nll_bottom){\includegraphics[width=\ablationImSize,keepaspectratio]{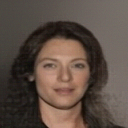}};
    
    \node[anchor=north] at ($(no_nll_top.north)+(0mm,5mm)$) {$\cancel{A_{\theta_T}}$};
    
    \draw[line width=0.5pt] ($(no_forward_kl_top.north)!0.5!(no_nll_top.north)+(0mm,\ablationLineTop)$) -- ($(no_forward_kl_bottom.south)!0.5!(no_nll_bottom.south)+(0mm,1mm)$);
    
    \node[right=\ablationRightGrid of no_nll_top,anchor=east] (no_remember_top){\includegraphics[width=\ablationImSize,keepaspectratio]{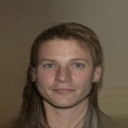}};
    
    \node[below=\ablationBelowGrid of no_remember_top,anchor=south] (no_remember_bottom){\includegraphics[width=\ablationImSize,keepaspectratio]{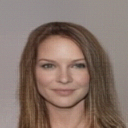}};
    
    \node[anchor=north] at ($(no_remember_top.north)+(0mm,5mm)$) {$\cancel{\mathcal{L}_{KL_F} + \mathcal{L}_{KL_R}}$};
    
    \draw[line width=0.5pt] ($(no_nll_top.north)!0.5!(no_remember_top.north)+(0mm,\ablationLineTop)$) -- ($(no_nll_bottom.south)!0.5!(no_remember_bottom.south)+(0mm,1mm)$);
    
    \node[right=\ablationRightGrid of no_remember_top,anchor=east] (trained_top){\includegraphics[width=\ablationImSize,keepaspectratio]{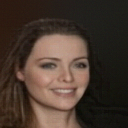}};
    
    \node[below=\ablationBelowGrid of trained_top,anchor=south] (trained_bottom){\includegraphics[width=\ablationImSize,keepaspectratio]{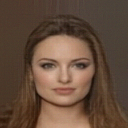}};
    
    \node[anchor=north] at ($(trained_top.north)+(0mm,4.5mm)$) {Tamed ($\theta_T$)};
    
    \draw[line width=0.5pt] ($(no_remember_top.north)!0.5!(trained_top.north)+(0mm,\ablationLineTop)$) -- ($(no_remember_bottom.south)!0.5!(trained_bottom.south)+(0mm,1mm)$);
    
    \end{scope}
    \end{tikzpicture}
    \caption{\textbf{Qualitative ablation comparison.} For a fixed couple of latent vectors drawn from the tractable prior distribution, the shown images were acquired by passing the latents through different normalizing flows. Each column represents the output of a different model. Each middle column represents a model that was trained while eliminating a different part of the loss, \eg $\cancel{\mathcal{L}_R}$ uses the entire loss term apart from the remember loss $\mathcal{L}_R$. The different loss components are presented in \cref{eq:remember_loss,eq:objective}.}
    \label{fig:ablation_qualitative}
\end{figure*}
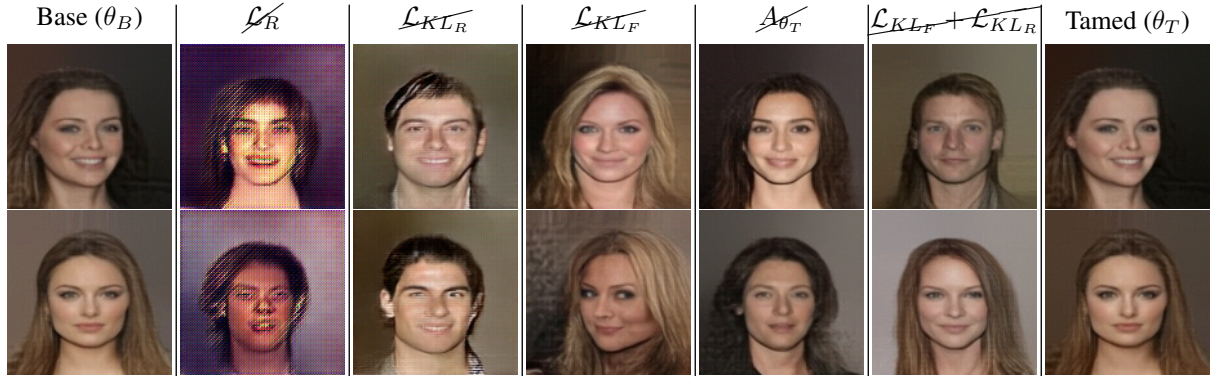
\cref{fig:ablation_qualitative} shows the results.

Without any loss that preserves the knowledge of the original objective (the $\mathcal{L}_R$ loss), the quality of the generated images is significantly worse.
Furthermore, we see that the reverse KL divergence loss, $\mathcal{L}_{KL_R}$, is vital to produce images with high quality. There are some parts of the total loss term that seem to have a lower effect on the generation quality (columns 4--6 in \cref{fig:ablation_qualitative}), but only when using the full loss objective, we get a model that preserves the original images with high quality.
This is evident from the figure, showing that only the rightmost column preserves the images that were generated using $\theta_B$.
This strengthens the assumption that taming preserves the latent space structure, as discussed in \cref{sec:forget_attribute}, and demonstrated in \cref{fig:const_latents}.

\textbf{Quantitative comparison.}\label{subsec:ablation_qunatitative}
\cref{tab:ablation_quantitatively} shows a comparison of the different ablated models. A full comparison can be found in the Supplementary. 
We see that regarding the NLL (measured in BPD in the table) when we use the KL divergence loss, reverse KL divergence is crucial to ensure a high likelihood of the training data, as it is suited for data generation tasks. When omitting $\mathcal{L}_R$ the FID~\cite{fid} score grows (worse), and the forgetting objective (the likelihood quantile column) grows as well. The growth in FID score means we drift from the original data, and the large likelihood score means the model does not forget the data it was supposed to forget. Omitting the $\mathcal{L}_{KL_{R}}$ term maintains similarity with the original data (low FID), but fails to forget (again, the likelihood is quite high).

\subsection{Limitations}
Our method is demonstrated only on normalizing flows, and we do not demonstrate it on explicit generative models at all. Moreover, our precise evaluation of forgetting data samples is based on the assumption of normality on the model's NLL training set. 
Although we do offer a qualitative measurement using a normality test as mentioned in \cref{sec:eval_forget}, models that do not align with this assumption will find the threshold we use less powerful and accurate.
\begin{table}[t]
    \centering
    \begin{adjustbox}{width=1.0\linewidth}
    \begin{tabular}{@{} ccccc @{}}
    \toprule
    Model&{BPD mean $(\downarrow)$}&{FID$(\downarrow)$}&
    {\makecell{Forget\\ threshold}}&
    {\makecell{Likelihood\\quantile $\q_{\theta}(\D_F)$}}\tabularnewline
    \midrule
    {Base ($\theta_B$)}&\hspace{2mm}$1.021$&$140.89$&-&-\tabularnewline
    {$\cancel{\mathcal{L}_R}$}&\hspace{0.3mm}$25.640$&$181.26$&\redX&
    $0.34$\tabularnewline
    $\cancel{\mathcal{L}_{KL_R}}$&\hspace{2mm}$2.254$& $\bm{110.34}$&\redX&$0.73$
    \tabularnewline
    {$\cancel{\mathcal{L}_{KL_F}}$}&\hspace{2mm}$1.072$&$121.45$&{\greencheck}&
    {\hspace{6.2mm}$2.45$e$-5$}\tabularnewline
    {$\cancel{\A_{\theta_T}}$}&\hspace{2mm}$1.025$&$141.99$&{\greencheck}&{\hspace{6.2mm}$1.02$e$-5$}
    \tabularnewline
    {$\cancel{\mathcal{L}_{KL_F}+\mathcal{L}_{KL_R}}$}&\hspace{2mm}$1.028$&$143.16$&{\greencheck}&{\hspace{6.2mm}$8.53$e$-6$}\tabularnewline
    Tamed ($\theta_T$)&\hspace{2mm}$\bm{1.020}$
    &$141.63$&
    {\greencheck}&
    {\hspace{6.2mm}$2.35$e$-5$}\tabularnewline
    \bottomrule
    \end{tabular}
    \end{adjustbox}
    \caption{\textbf{Quantitative ablation study.} Removing parts of our loss affects the NLL of $\D_R$ (in BPD~\cite{papamakarios2017masked} units), the generation quality (FID), and the forgetness of $\D_F$, in terms achieving the threshold (\cref{eq:forget_threshold}) and the likelihood quantile $q_\theta(\D_F)$. For models names notation see \cref{fig:ablation_qualitative,eq:remember_loss,eq:objective}.}
    \label{tab:ablation_quantitatively}
    \vspace{-3mm}
\end{table}

\section{Ethical considerations}%
As generative models have gained immense popularity in recent years, the wider public interest in these models increased and important issues arise with respect to the generation of hateful, fake, explicit, and biased content. Many entities that train these models are concerned about the potential risks and opt out of their public release. In our work, we try to proceed in a direction that can help moderate these malicious applications, and help to mitigate them.
Although our method can be used in a positive manner, unfortunately, it can also be used in the opposite direction, \eg, increasing a known bias of a model instead of debiasing it.
We are aware of these potential risks and suggest that individuals using this work will do so carefully, realizing that it can be exploited in the wrong hands.
\section{Conclusion}
In this work, we proposed an approach towards taming normalizing flow models, controlling their output by increasing or decreasing the probability of generating specific data. We demonstrated different aspects of taming on human faces, showing how to change generation probability both locally and globally, in a fast procedure that supports the high scale usage of generative models. Taming provides an easy modification tool, with minimal collateral damage to the model. 
Although taming is demonstrated only on normalizing flows, our approach can be extended to other generative models based on exact likelihood estimation.

\section{Acknowledgments}
We would like to thank Itay Evron and Ori Katz for their valuable constructive feedback. This work was supported in part by the Israel Science Foundation (grant No. 1574/21).
{\small
\bibliographystyle{ieee_fullname}
\bibliography{egbib}
}

\appendix
\clearpage
\arxiv{\section*{\LARGE Appendices}}{}

\renewcommand\thefigure{\thesection.\arabic{figure}}
\renewcommand\thetable{\thesection.\arabic{table}}
\setcounter{figure}{0}
\setcounter{table}{0}

In the next sections, we provide additional details, results, 
and visualizations, further demonstrating our method's applications.
\section{Additional details}\label{supp:sec:additional_details}
First, we elaborate on technical details regarding the implementation of our method, as explained in \cref{sec:experiments}.
We trained a Glow~\cite{glow} base model to produce RGB images with dimensions $128\times128$. The training was done for 590K iterations with a batch size of 32, for a total of $316.3$ hours, using $4$ $12$GB Titan Xp GPUs. The model has $4$ blocks of $32$ flows, each consisting of activation normalization layers, $1\times1$ LU decomposed convolution and additive coupling. The model is trained using an Adam~\cite{adam} optimizer with learning rate $5\cdot10^{-5}$ and betas $(\beta_1, \beta_2) = (0.9, 0.999)$. Images are quantized to $5$ bits and learned using the continuous dequantization process as done in previous work~\cite{glow,gen_models_eval_dequnatization}. Since the dequantization introduces the addition of random noise proportional to the size of quantization bins, every likelihood estimation we perform in \cref{subsec:forget_identity_train_set} is averaged over 10 estimations using different random noise. For the forgetting process, we use a threshold of $\delta=4$ and a bound of $\epsilon=0.15\cdot\delta$. We use the hyperparameters $\alpha=0.6$ and $\gamma=0.6$ in all our experiments, chosen using a grid search.
As we trained $\theta_B$ on the training set of CelebA~\cite{celeba}, we used the validation set of CelebA as the holdout set in this evaluation and all upcoming demonstrations, unless specified otherwise.

In \cref{tab:forget_identity}, each experiment is averaged over 5 experiments with different identities. The nearest neighbors are chosen using the 5 nearest neighbors, selected using the average ``Cosine Distance'' between the ArcFace~\cite{deng2019arcface} face embeddings.

The tamed model used for \cref{fig:generated_samples} is a model that was trained to forget 15 images of an identity, similar to the last row in \cref{tab:forget_identity}.

The classifier used in \cref{fig:forget_multiple_attributes} was trained on the attributes of CelebA~\cite{celeba}, using a ResNet50~\cite{he2016deep} backbone and achieving an AUC $> 0.99$ for every binary attribute in CelebA on a holdout set.

\begin{figure*}[t]
\centering
\begin{tikzpicture}
\node at (0,0) (train_gaussian) {\includegraphics[width=0.47\linewidth]{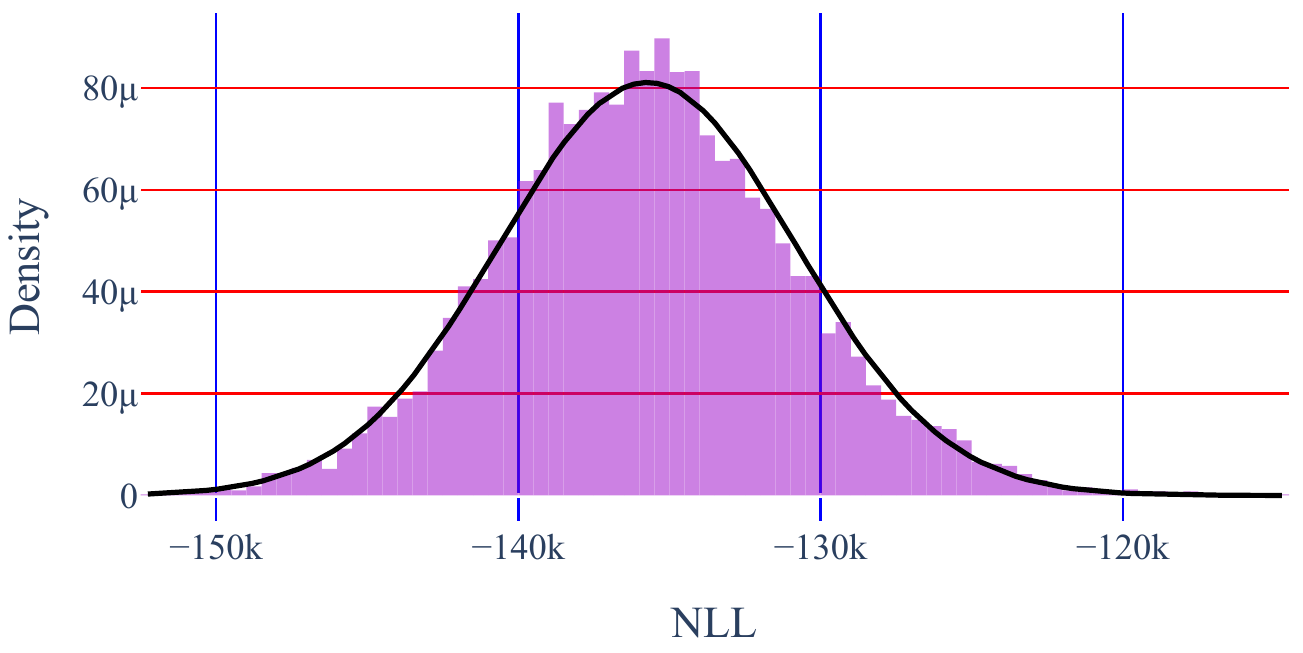}};

\node[right=1mm of train_gaussian] (valid_gaussian)
{\includegraphics[width=0.47\linewidth]{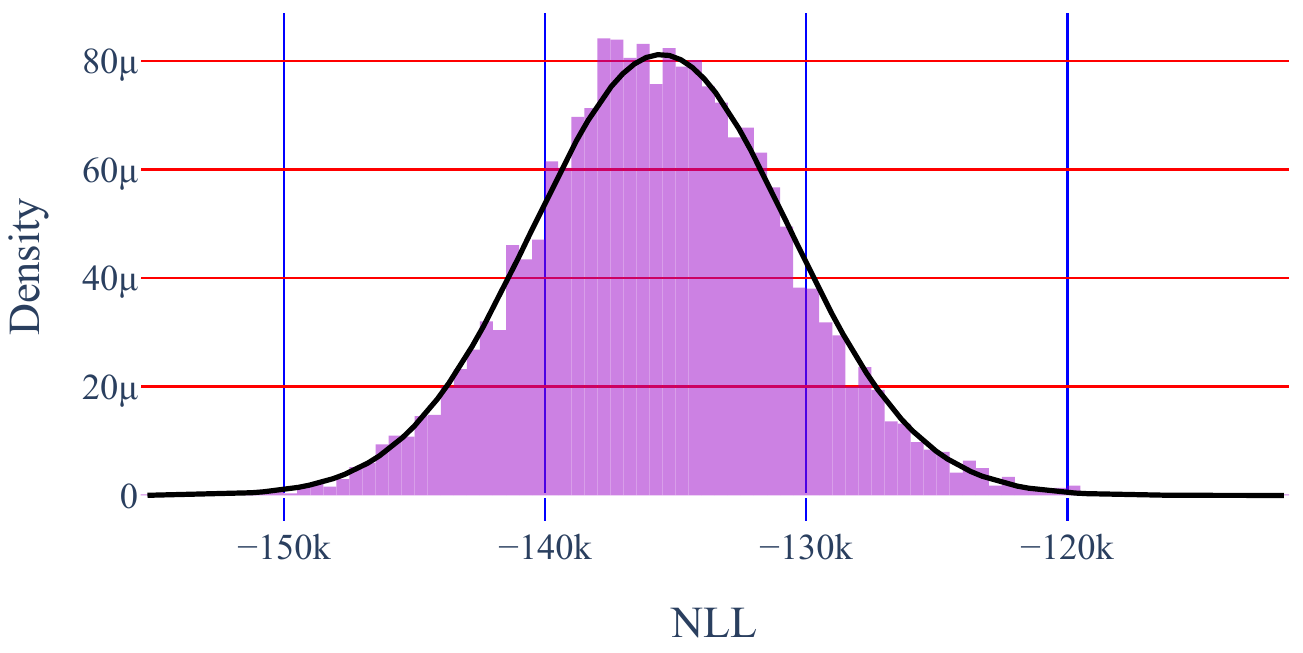}};

\node[below=1mm of train_gaussian] (train_qq) 
{\includegraphics[width=0.47\linewidth]{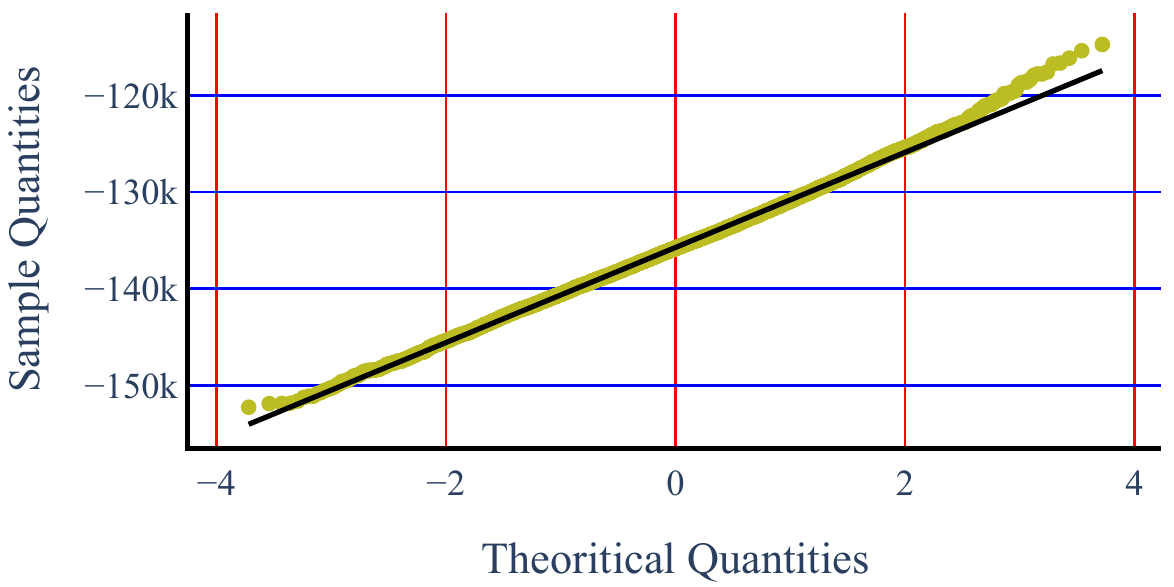}};

\node[below=1mm of valid_gaussian] (valid_qq) 
{\includegraphics[width=0.47\linewidth]{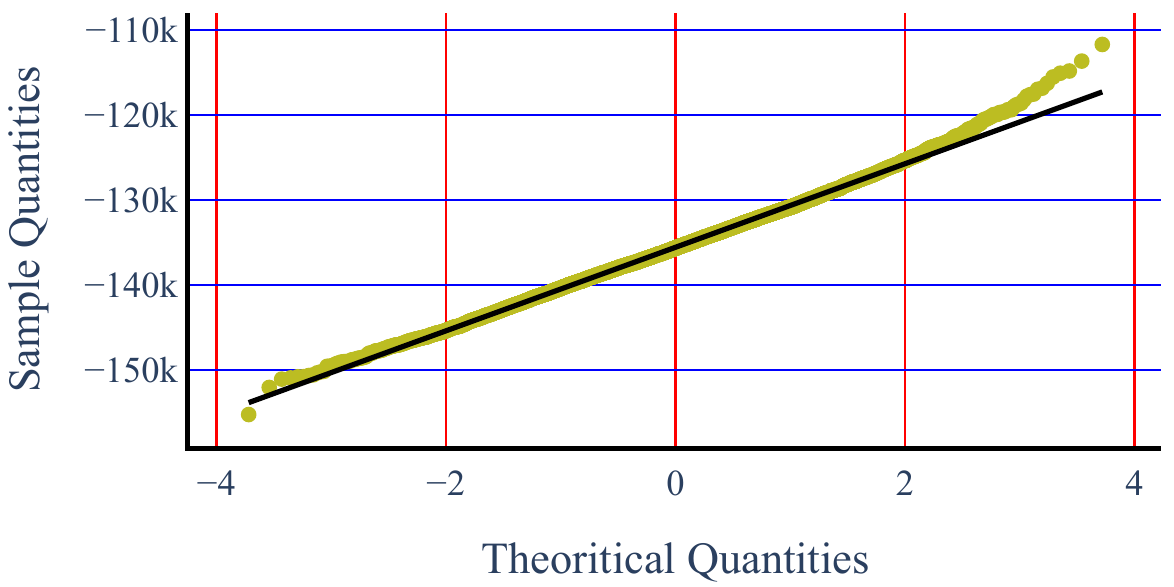}};

\node[above=5mm of train_gaussian] {Train set};
\node[above=5mm of valid_gaussian] {Holdout set};
\end{tikzpicture}
\caption{\textbf{Base model ($\theta_B$) NLL normal assumption.} For the base model's training set (left) and a similar holdout set (right), we show a QQ-plot against normal distribution (lower row). We also show (upper row) the normalized density histogram (purple) and a Gaussian estimation (black line) of the distribution. These results suggest that a normal distribution assumption fits this case.}
\label{supp:fig:base_distribution}
\end{figure*}

Next, we discuss the normality assumption as explained in \cref{subsec:task}. We assumed the NLL distribution of the base model on the training data is normal. To support this assumption, \cref{supp:fig:base_distribution} visually compares the distribution with a normal estimation, along with \mbox{QQ-plots} that further support this claim.
We also performed a \textit{Kolmogorov–Smirnov test}~\cite{ks_test} and received a p-value of $0.95,0.54$ on 2000 random samples of CelebA's training and validation sets, respectively. Thus, these results suggest that assuming a normal distribution is 
reasonable.

\setcounter{figure}{0}
\setcounter{table}{0}
\section{Results}\label{supp:sec:results}
In this section, we discuss additional results associated with experiments from our paper. We show:
\begin{enumerate}[label=\upshape(\Roman*),ref= (\Roman*)]
    \item\label{supp:results:item:identity}  Additional scenarios and details for the experiment of ``Taming an attribute'' (\cref{subsec:forget_identity_train_set}).
    \item\label{supp:results:item:remember_set} Results evaluating our method on the remember set $\D_R$.
    \item\label{supp:results:item:without_train_set}  A full comparison of the experiment of ``Taming without the training set'' (\cref{sec:forget_without_train_set}).
\end{enumerate}

We first discuss the experiment in \cref{subsec:forget_identity_train_set} (\cref{supp:results:item:identity}).
\cref{supp:tab:raw_delta_4,supp:tab:raw_delta_3,supp:tab:raw_delta_2,supp:tab:out_of_training} include a more detailed analysis of \cref{tab:forget_identity}, with additional details regarding the likelihood quantiles and the running time. 

For example, the first row in  \cref{supp:tab:raw_delta_4} shows that when forgetting 1 image, we are able to reach the forget threshold. Regarding the forget set $\D_F$, the likelihood quantile of the base model is $0.47$ while for the tamed model it is $< 10^{-3}$, resulting in a quantile drop of $0.47$.
This row also shows that the running time for this experiment is 3.2 minutes.

\cref{supp:tab:raw_delta_3,supp:tab:raw_delta_2} include the results for using a different forget threshold, $\delta=2$ and $\delta=3$ respectively.
\cref{supp:tab:out_of_training} includes results for using a forget set outside of the training set, \ie, $\D_F\not\subset\D_R$. The images in $\D_F$ are all from an identity of a holdout set from the same distribution. 

These tables show that even for the aforementioned different settings, we are able to forget the identity ($\D_F$) while reducing the likelihood of a holdout set of its images ($\D_F'$), with marginal impact on the remember distribution ($\D_R$, $\D_R^{id}$ and $\D_R^{NN}$).

Now we turn to inspect whether the time to forget an identity depends on the number of images the model was trained on. To do so, we trained an additional base model just on CelebA. This model was trained on $162{,}770$ images, while the original one, trained additionally on FFHQ~\cite{stylegan}, was trained on $232{,}770$. For both models, training stopped with the same performance (in NLL) on CelebA's training set. 
In \cref{tab:forget_identity_timing}, we compare the running time of these models and see that even for a smaller training set the running time is comparable and fast.

As discussed in \cref{sec:ablation}, in \cref{supp:fig:ablation_distirbution} we compare the distribution of NLL values on the training set of the base model, for different ablated models. Some models are not shown in the figure, as they have a distribution that is visually indistinguishable from the shown distributions of the base and tamed models. The figure shows that without using the forward KL divergence loss ($\cancel{\mathcal{L}_{KL_F}}$), the distribution is worse, but it's also more ``narrow'', fitting the mode-seeking behavior of the reverse KL divergence. On the contrary, without the reverse KL divergence ($\cancel{\mathcal{L}_{KL_R}}$), which is known to be important for generative tasks, the performance is bad, and fits the mean-seeking behavior of forward kl divergence, attempting to cover more regions.

Next, we discuss \cref{supp:results:item:remember_set}, showing how our method preserves the NLL distribution of the remember set $\D_R$. In \cref{sec:experiments}, we showed results focusing on the forget set $\D_F$. We now show results, focusing on $\D_R$. This is demonstrated by showing this distribution before and after taming, as seen in \cref{supp:fig:preserve_dist}. The figure visualizes the differences between the distributions of the base model ($\theta_B$) and the tamed model ($\theta_T$), for both the training set and a holdout set. This is done using the normalized density histogram of these distributions, and also by estimating the parameters of a normal distribution using the distributions' observations.
The distribution pairs in \cref{supp:fig:preserve_dist} are all similar, indicating that we successfully forget the target(s), without heavily impacting the rest of the distribution.

\begin{figure}[t]
    \centering
    \begin{tikzpicture}
    \def\arrowLift#1{\raisebox{0.4ex}}
    \node [] (fig)
    {\includegraphics[width=\linewidth]{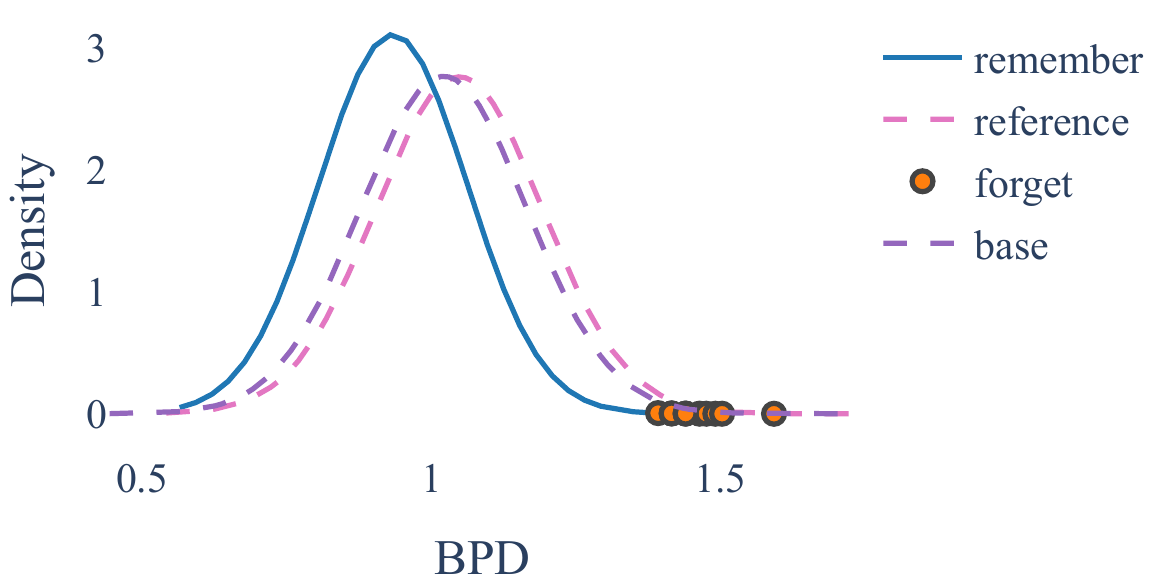}};

    \draw [->,line width=1.6pt,postaction={decorate,decoration={text along path,text align=center,text={|\small \arrowLift| More likely},reverse path}}] ($(fig.south)+(0mm,11mm)$) -- ($(fig.south)+(-20mm,11mm)$);
    
    \end{tikzpicture}
    \caption{\textbf{Forget without training data access.} Full comparison of \cref{fig:forget_without_trainset}. Here we see not only the NLL distributions of $\theta_T$ on the remember set (solid), but also on the training set $\D$ (dashed pink). The distribution of the base model $\theta_B$ on the training data (dashed purple) is close to the tamed model on the same data, while the forget images (orange dots) reach the forget threshold.}

    \label{supp:fig:forget_without_trainset}
\end{figure}

Lastly, we discuss the experiment in \cref{sec:forget_without_train_set} (\cref{supp:results:item:without_train_set}).
\cref{supp:fig:forget_without_trainset} shows a more detailed comparison of \cref{fig:forget_without_trainset}, additionally showing the NLL distribution of the tamed model ($\theta_T$) on the original training data. 
We see that while there is some decrease in the likelihood of the original training data, this change is much smaller than the difference between the original training data $\D$ and the remember set $\D_R$, \ie, $\{-\log\p_{\theta_T}(\D)\}$ and $\{-\log\p_{\theta_T}(\D_R)\}$, respectively.

\begin{figure}
    \centering
    \includegraphics[width=\linewidth]{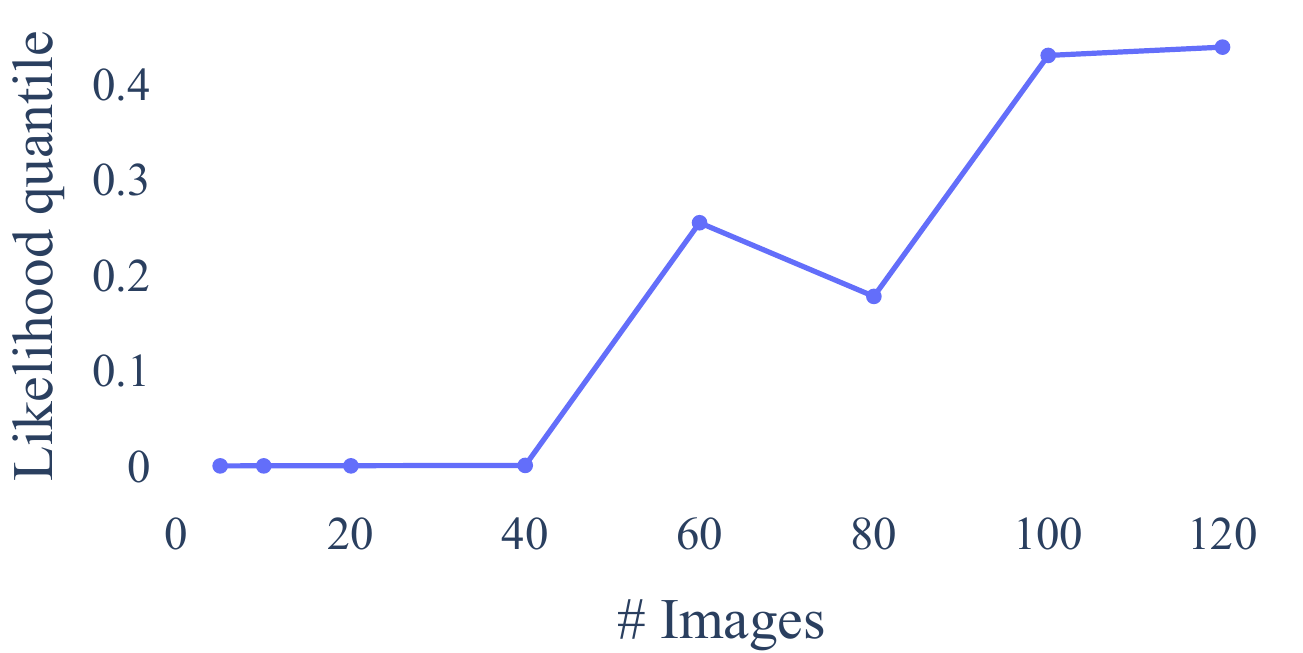}
    \caption{\textbf{Taming success against number of samples.}
    The plot shows the likelihood quantile of different models without any access to the training data. $\D_R$ contains 1000 images out of the base model's training set. Each model is tamed to forget a different number of images (the x-axis).
    We see that when $\D_F$ is small, taming can yield a big likelihood reduction, while on a lot of images the average difference drops.}
    \label{fig:no_data_acess}
    \vspace{-3mm}
\end{figure}

In \cref{fig:no_data_acess}, We evaluate the impact of the forget set size ($\abs{\D_F}$) on our method, \wrt the experiment in \cref{sec:forget_without_train_set}.
As the figure shows, when the size of $\D_F$ is small (\ie, $|\D_F|<40$) the average likelihood quantile remains near zero.  When $|\D_F|>40$, the average likelihood quantile increases. This is aligned with the different settings of our method, as we showed in \cref{sec:forget_attribute} where we used larger sets of forget images $\D_F$.

\begin{table*}[t]
    \centering
   \begin{subtable}{1.0\linewidth}
     \begin{adjustbox}{width=1.0\linewidth}
    \begin{tabular}{@{}cccccccccccccccccc@{}}
    \toprule
    \multirow{2}{*}{\# Images}&\multirow{2}{*}{\makecell{Forget\\threshold}}&\multicolumn{3}{c}{Forget set $\D_F$}&\multicolumn{3}{c}{Forget reference set $\D'_F$}&\multicolumn{3}{c}{Remember set $\D_R$ }&\multicolumn{3}{c}{Unseen identities $\D^{\textrm{id}}_R$}&\multicolumn{3}{c}{Nearest identities $\D_R^{\textrm{NN}}$}&\multirow{2}{*}{Time[minutes]}{}\tabularnewline
    
    \cmidrule(l{2pt}r{2pt}){3-5} \cmidrule(l{2pt}r{2pt}){6-8} \cmidrule(l{2pt}r{2pt}){9-11} \cmidrule(l{2pt}r{2pt}){12-14} \cmidrule(l{2pt}r{2pt}){15-17}
    
    &&$\q_{\theta_B}(\cdot)$&$\q_{\theta_T}(\cdot)$&$\QD(\cdot)$&$\q_{\theta_B}(\cdot)$&$\q_{\theta_T}(\cdot)$&$\QD(\cdot)$&$\q_{\theta_B}(\cdot)$&$\q_{\theta_T}(\cdot)$&$\QD(\cdot)$&$\q_{\theta_B}(\cdot)$&$\q_{\theta_T}(\cdot)$&$\QD(\cdot)$&$\q_{\theta_B}(\cdot)$&$\q_{\theta_T}(\cdot)$&$\QD(\cdot)$\tabularnewline
    \midrule
    
    1&\greencheck&$0.47$&$<10^{-3}$&$0.47$&$0.48$&$0.47$&$0.01$&$0.45$&$0.44$&$<10^{-2}$&$0.36$&$0.36$&$<10^{-3}$&$0.52$&$0.51$&$0.01$&3.2\tabularnewline
    4&\greencheck&$0.41$&$<10^{-4}$&$0.41$&$0.48$&$0.41$&$0.07$&$0.45$&$0.44$&$<10^{-2}$&$0.36$&$0.36$&$0.01$&$0.52$&$0.49$&$0.03$&9.3\tabularnewline
    8&\greencheck&$0.32$&$<10^{-4}$&$0.32$&$0.48$&$0.39$&$0.08$&$0.45$&$0.45$&$<10^{-3}$&$0.36$&$0.36$&$<10^{-2}$&$0.52$&$0.51$&$0.01$&16.2\tabularnewline
    15&\greencheck&$0.37$&$<10^{-4}$&$0.37$&$0.48$&$0.33$&$0.15$&$0.45$&$0.45$&$<10^{-3}$&$0.36$&$0.36$&$<10^{-2}$&$0.52$&$0.52$&$0.01$&17.6\tabularnewline
    
    \bottomrule
  \end{tabular}

  \end{adjustbox}

    \caption{Extensive results for the experiment in \cref{tab:forget_identity}.}
    \label{supp:tab:raw_delta_4}
   \end{subtable}

    \begin{subtable}{1.0\linewidth}
      \begin{adjustbox}{width=1.0\linewidth}
    \begin{tabular}{@{}cccccccccccccccccc@{}}
    \toprule
    \multirow{2}{*}{\# Images}&\multirow{2}{*}{\makecell{Forget\\threshold}}&\multicolumn{3}{c}{Forget set $\D_F$}&\multicolumn{3}{c}{Forget reference set $\D'_F$}&\multicolumn{3}{c}{Remember set $\D_R$ }&\multicolumn{3}{c}{Unseen identities $\D^{\textrm{id}}_R$}&\multicolumn{3}{c}{Nearest identities $\D_R^{\textrm{NN}}$}&\multirow{2}{*}{Time[minutes]}{}\tabularnewline
    
    \cmidrule(l{2pt}r{2pt}){3-5} \cmidrule(l{2pt}r{2pt}){6-8} \cmidrule(l{2pt}r{2pt}){9-11} \cmidrule(l{2pt}r{2pt}){12-14} \cmidrule(l{2pt}r{2pt}){15-17}
    
    &&$\q_{\theta_B}(\cdot)$&$\q_{\theta_T}(\cdot)$&$\QD(\cdot)$&$\q_{\theta_B}(\cdot)$&$\q_{\theta_T}(\cdot)$&$\QD(\cdot)$&$\q_{\theta_B}(\cdot)$&$\q_{\theta_T}(\cdot)$&$\QD(\cdot)$&$\q_{\theta_B}(\cdot)$&$\q_{\theta_T}(\cdot)$&$\QD(\cdot)$&$\q_{\theta_B}(\cdot)$&$\q_{\theta_T}(\cdot)$&$\QD(\cdot)$\tabularnewline
    \midrule
    1&\greencheck&$0.47$&$<10^{-2}$&$0.46$&$0.48$&$0.47$&$0.01$&$0.51$&$0.51$&$<10^{-3}$&$0.36$&$0.36$&$<10^{-2}$&$0.52$&$0.52$&$0.01$&4.2\tabularnewline
    4&\greencheck&$0.41$&$<10^{-2}$&$0.41$&$0.48$&$0.44$&$0.04$&$0.50$&$0.50$&$<10^{-3}$&$0.36$&$0.36$&$<10^{-2}$&$0.52$&$0.51$&$0.01$&12.4\tabularnewline
    8&\greencheck&$0.32$&$<10^{-2}$&$0.32$&$0.48$&$0.42$&$0.06$&$0.49$&$0.49$&$<10^{-2}$&$0.36$&$0.36$&$<10^{-2}$&$0.52$&$0.51$&$0.01$&19.1\tabularnewline
    15&\greencheck&$0.37$&$<10^{-2}$&$0.37$&$0.48$&$0.37$&$0.11$&$0.49$&$0.49$&$<10^{-2}$&$0.36$&$0.36$&$<10^{-3}$&$0.52$&$0.52$&$0.01$&22.3\tabularnewline
    
    \bottomrule
  \end{tabular}

  \end{adjustbox}
    
    \caption{Results for a different forget threshold, $\delta=3$.}
    \label{supp:tab:raw_delta_3}
   \end{subtable}

    \begin{subtable}{1.0\linewidth}
      \begin{adjustbox}{width=1.0\linewidth}
    \begin{tabular}{@{}cccccccccccccccccc@{}}
    \toprule
    \multirow{2}{*}{\# Images}&\multirow{2}{*}{\makecell{Forget\\threshold}}&\multicolumn{3}{c}{Forget set $\D_F$}&\multicolumn{3}{c}{Forget reference set $\D'_F$}&\multicolumn{3}{c}{Remember set $\D_R$ }&\multicolumn{3}{c}{Unseen identities $\D^{\textrm{id}}_R$}&\multicolumn{3}{c}{Nearest identities $\D_R^{\textrm{NN}}$}&\multirow{2}{*}{Time[minutes]}{}\tabularnewline
    
    \cmidrule(l{2pt}r{2pt}){3-5} \cmidrule(l{2pt}r{2pt}){6-8} \cmidrule(l{2pt}r{2pt}){9-11} \cmidrule(l{2pt}r{2pt}){12-14} \cmidrule(l{2pt}r{2pt}){15-17}
    
    &&$\q_{\theta_B}(\cdot)$&$\q_{\theta_T}(\cdot)$&$\QD(\cdot)$&$\q_{\theta_B}(\cdot)$&$\q_{\theta_T}(\cdot)$&$\QD(\cdot)$&$\q_{\theta_B}(\cdot)$&$\q_{\theta_T}(\cdot)$&$\QD(\cdot)$&$\q_{\theta_B}(\cdot)$&$\q_{\theta_T}(\cdot)$&$\QD(\cdot)$&$\q_{\theta_B}(\cdot)$&$\q_{\theta_T}(\cdot)$&$\QD(\cdot)$\tabularnewline
    \midrule
    
    1&\greencheck&$0.47$&$0.04$&$0.43$&$0.48$&$0.47$&$0.01$&$0.45$&$0.45$&$<10^{-3}$&$0.36$&$0.36$&$<10^{-3}$&$0.52$&$0.52$&$0.01$&6.1\tabularnewline
    4&\greencheck&$0.41$&$0.03$&$0.38$&$0.48$&$0.46$&$0.02$&$0.49$&$0.49$&$<10^{-2}$&$0.36$&$0.36$&$<10^{-2}$&$0.52$&$0.52$&$<10^{-2}$&22.4\tabularnewline
    8&\greencheck&$0.32$&$0.02$&$0.30$&$0.48$&$0.45$&$0.02$&$0.51$&$0.50$&$<10^{-2}$&$0.36$&$0.36$&$<10^{-2}$&$0.52$&$0.52$&$0.01$&29.8\tabularnewline
    15&\greencheck&$0.37$&$0.02$&$0.35$&$0.48$&$0.43$&$0.05$&$0.50$&$0.50$&$<10^{-2}$&$0.36$&$0.36$&$<10^{-2}$&$0.52$&$0.52$&$0.01$&36.0\tabularnewline
    
    \bottomrule
  \end{tabular}

  \end{adjustbox}
    
    \caption{Results for a different forget threshold, $\delta=2$.}
    \label{supp:tab:raw_delta_2}
   \end{subtable}

    \begin{subtable}{1.0\linewidth}
      \begin{adjustbox}{width=1.0\linewidth}
    \begin{tabular}{@{}cccccccccccccccccc@{}}
    \toprule
    \multirow{2}{*}{\# Images}&\multirow{2}{*}{\makecell{Forget\\threshold}}&\multicolumn{3}{c}{Forget set $\D_F$}&\multicolumn{3}{c}{Forget reference set $\D'_F$}&\multicolumn{3}{c}{Remember set $\D_R$ }&\multicolumn{3}{c}{Unseen identities $\D^{\textrm{id}}_R$}&\multicolumn{3}{c}{Nearest identities $\D_R^{\textrm{NN}}$}&\multirow{2}{*}{Time[minutes]}{}\tabularnewline
    
    \cmidrule(l{2pt}r{2pt}){3-5} \cmidrule(l{2pt}r{2pt}){6-8} \cmidrule(l{2pt}r{2pt}){9-11} \cmidrule(l{2pt}r{2pt}){12-14} \cmidrule(l{2pt}r{2pt}){15-17}
    
    &&$\q_{\theta_B}(\cdot)$&$\q_{\theta_T}(\cdot)$&$\QD(\cdot)$&$\q_{\theta_B}(\cdot)$&$\q_{\theta_T}(\cdot)$&$\QD(\cdot)$&$\q_{\theta_B}(\cdot)$&$\q_{\theta_T}(\cdot)$&$\QD(\cdot)$&$\q_{\theta_B}(\cdot)$&$\q_{\theta_T}(\cdot)$&$\QD(\cdot)$&$\q_{\theta_B}(\cdot)$&$\q_{\theta_T}(\cdot)$&$\QD(\cdot)$\tabularnewline
    \midrule
    
    1&\greencheck&$0.48$&$<10^{-4}$&$0.48$&$0.32$&$0.28$&$0.04$&$0.50$&$0.49$&$<10^{-2}$&$0.34$&$0.34$&$<10^{-2}$&$0.52$&$0.51$&$0.01$&3.7\tabularnewline
    4&\greencheck&$0.40$&$<10^{-4}$&$0.40$&$0.32$&$0.23$&$0.10$&$0.50$&$0.50$&$<10^{-2}$&$0.34$&$0.34$&$<10^{-2}$&$0.52$&$0.51$&$0.01$&8.7\tabularnewline
    8&\greencheck&$0.39$&$<10^{-4}$&$0.39$&$0.32$&$0.14$&$0.19$&$0.49$&$0.49$&$<10^{-2}$&$0.34$&$0.34$&$<10^{-2}$&$0.52$&$0.50$&$0.02$&12.7\tabularnewline
    15&\greencheck&$0.41$&$<10^{-4}$&$0.41$&$0.32$&$0.09$&$0.23$&$0.51$&$0.51$&$<10^{-2}$&$0.34$&$0.34$&$<10^{-3}$&$0.52$&$0.50$&$0.02$&21.5\tabularnewline
    
    \bottomrule
  \end{tabular}

  \end{adjustbox}

    \caption{Results on an identity outside the training set (from a holdout set of the same distribution).}
    \label{supp:tab:out_of_training}
   \end{subtable}
   
    \caption{\textbf{Forget an identity - Comprehensive evaluation.} Additional results for the experiment in \cref{tab:forget_identity}, including results for additional thresholds, and forgetting an identity outside the training set. These tables include the Quantile drop ($\QD_{\theta_B,\theta_T}(\cdot)$), along with the likelihood quantiles ($\q_\theta(\cdot)$), for different evaluated sets. It also includes the running time in minutes of every experiment.}
    \label{supp:tab:forget_identities_main}
\end{table*}
\begin{table}[t]
  \centering
    \begin{tabular}{@{}D{.}{.}{2.0}D{.}{.}{3.1}D{.}{.}{1.4}D{.}{.}{3.1}D{.}{.}{1.4}@{}}
    \toprule
    \multicolumn{1}{c}{\multirow{2}{*}{\# Images}}&\multicolumn{2}{c}{CelebA+FFHQ}&\multicolumn{2}{c}{CelebA}\tabularnewline
    \cmidrule(lr){2-3} \cmidrule(lr{2pt}){4-5}
    &\multicolumn{1}{c}{T[minutes]}&\multicolumn{1}{c}{T[\%]}&\multicolumn{1}{c}{T[minutes]}&\multicolumn{1}{c}{T[\%]}\tabularnewline
    \midrule
    1&3.2&0.02\%&3.9&0.04\%\tabularnewline
    4&9.3&0.06\%&9.0&0.09\%\tabularnewline
    8&16.2&0.09\%&16.5&0.16\%\tabularnewline
    15&17.6&0.15\%&21.9&0.21\%\tabularnewline
    \bottomrule
  \end{tabular}
  \vspace{2mm}
  \caption{\textbf{Training size effect on running time.} We compare the running time for taming an identity using two different base models ($\theta_B$), trained using different training set size ($\D$).
  T[minutes] is the time taken to run this experiment in minutes. T[\%] is the experiment's runtime divided by the base model’s total training time, in percentages.}
  \label{tab:forget_identity_timing}
\end{table}
\begin{figure}
    \centering
    \begin{tikzpicture}
    \node at (0,0) {\includegraphics[width=0.9739\linewidth,keepaspectratio]{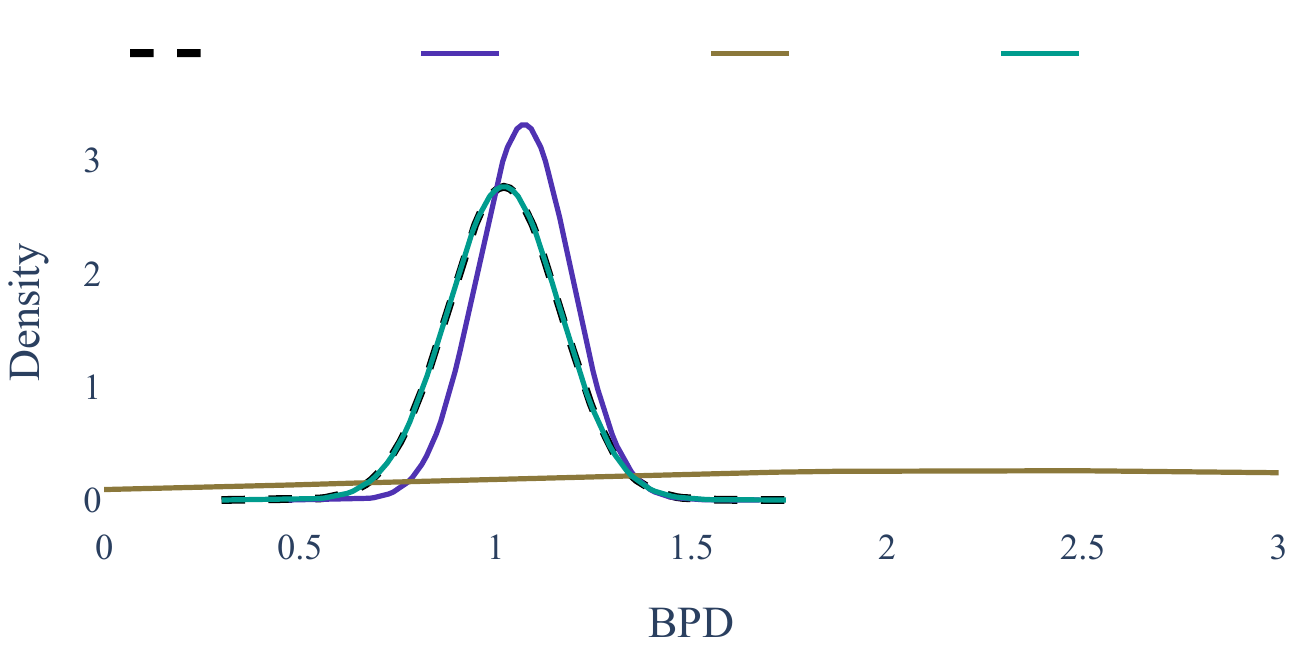}};
    \node at (-2.1,1.7) {{\scriptsize Base ($\theta_B$)}};
    \node at (-0.3,1.7) {{\scriptsize  $\cancel{\mathcal{L}_{KL_F}}$}};
    \node at (1.6,1.7) {{\scriptsize  $\cancel{\mathcal{L}_{KL_R}}$}};
    \node at (3.4,1.7) {{\scriptsize Tamed ($\theta_T$)}};
    \end{tikzpicture}
    
    \caption{\textbf{Ablation distribution comparison.} Comparison of the NLL distribution of models presented in \cref{sec:ablation}. Notice how $\cancel{\mathcal{L}_{KL_R}}$ and $\cancel{\mathcal{L}_{KL_R}}$ do not preserve the base distribution well.}
    \label{supp:fig:ablation_distirbution}
\end{figure}
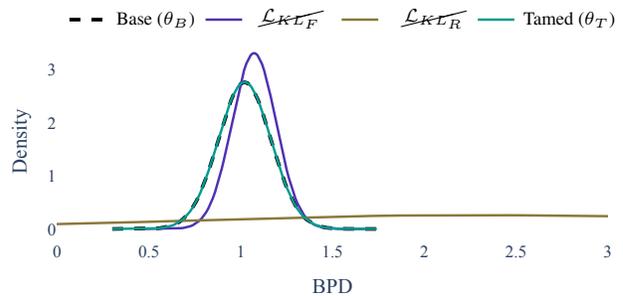
\begin{figure*}
\setlength{\tabcolsep}{0.5pt}
\renewcommand{\arraystretch}{0.455}
\newcommand{\cbox}[1]{\raisebox{\depth}{\fcolorbox{black}{#1}{\null}}}
\definecolor{dist_green}{RGB}{87,239,66}
\definecolor{dist_blue}{RGB}{39,56,196}
\definecolor{dist_pink}{RGB}{239,92,235}

\centering
\begin{tabular}{lcc}
 \# Images&Train set&Holdout set\tabularnewline
 \raisebox{23mm}1&\includegraphics[width=0.45\linewidth]{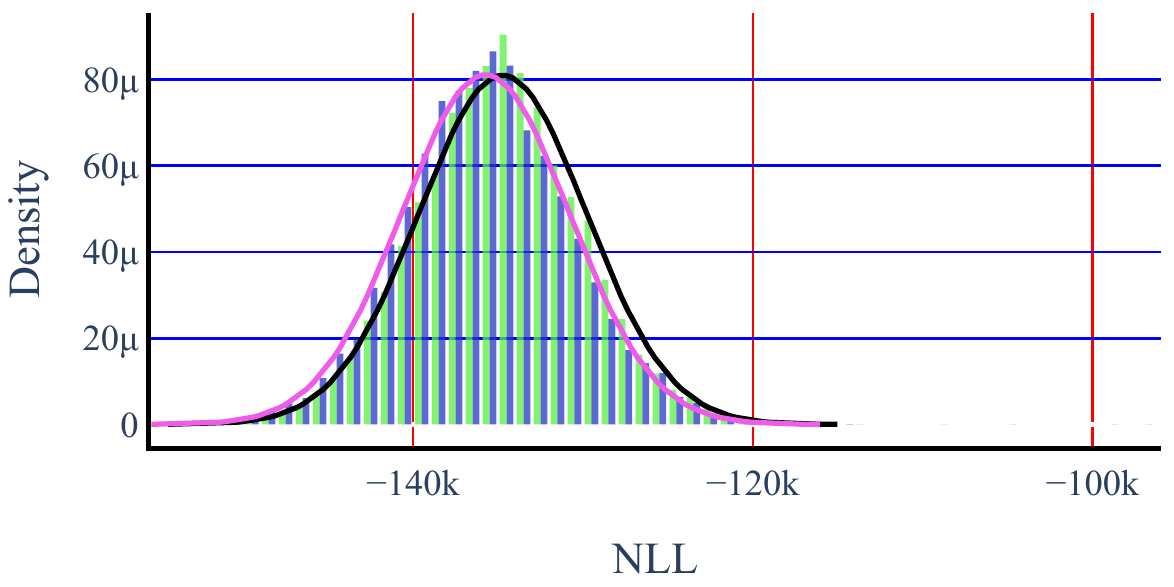} &
 \includegraphics[width=0.45\linewidth]{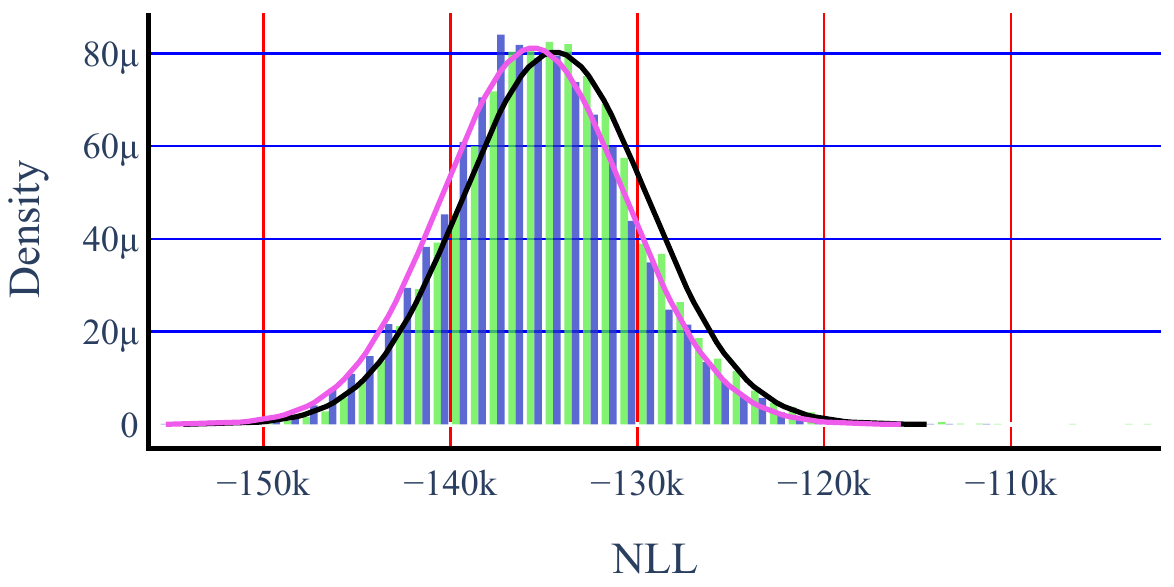}\tabularnewline
 \raisebox{23mm}4&\includegraphics[width=0.45\linewidth]{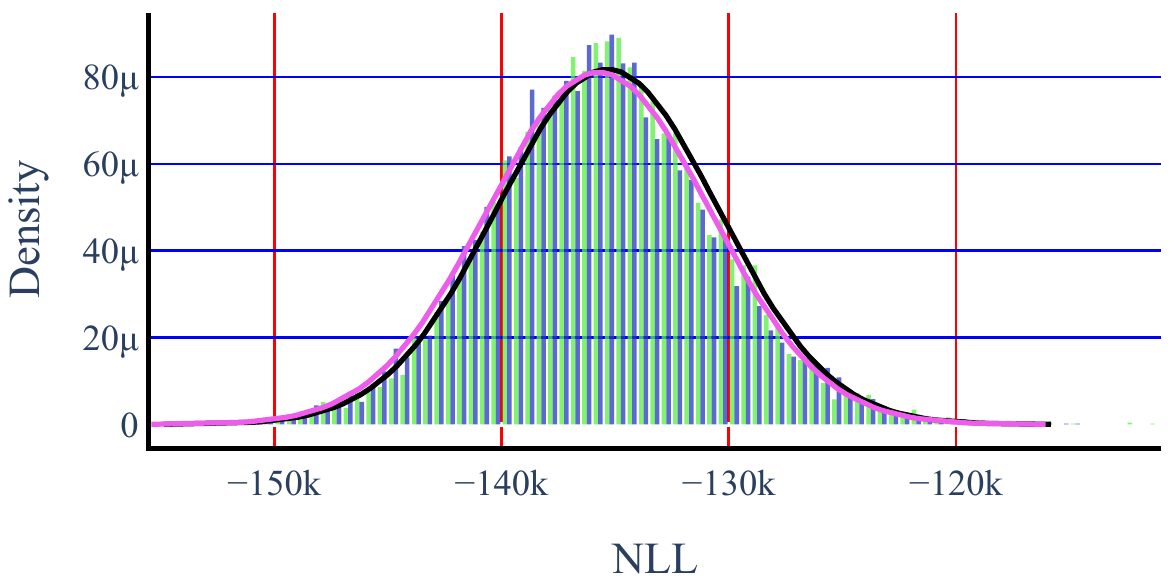} &
 \includegraphics[width=0.45\linewidth]{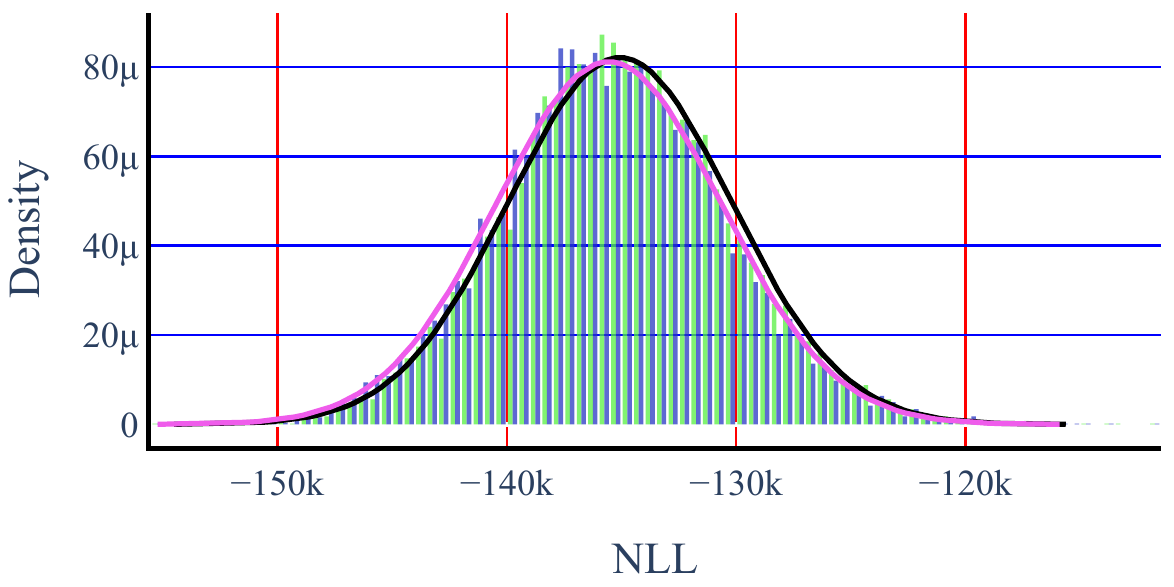}\tabularnewline
 \raisebox{23mm}8&\includegraphics[width=0.45\linewidth]{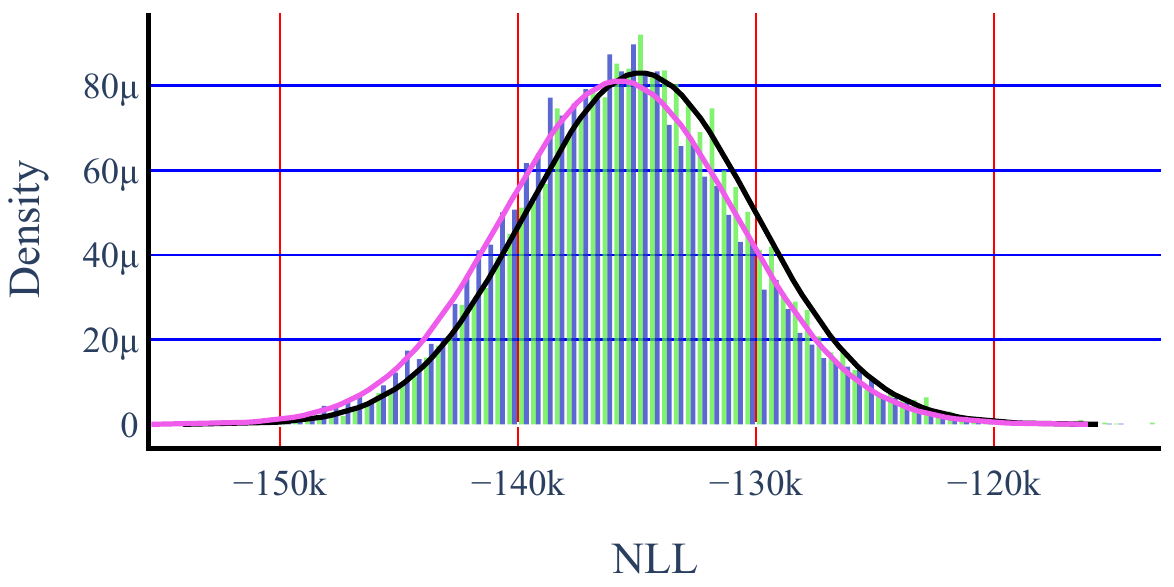} &
 \includegraphics[width=0.45\linewidth]{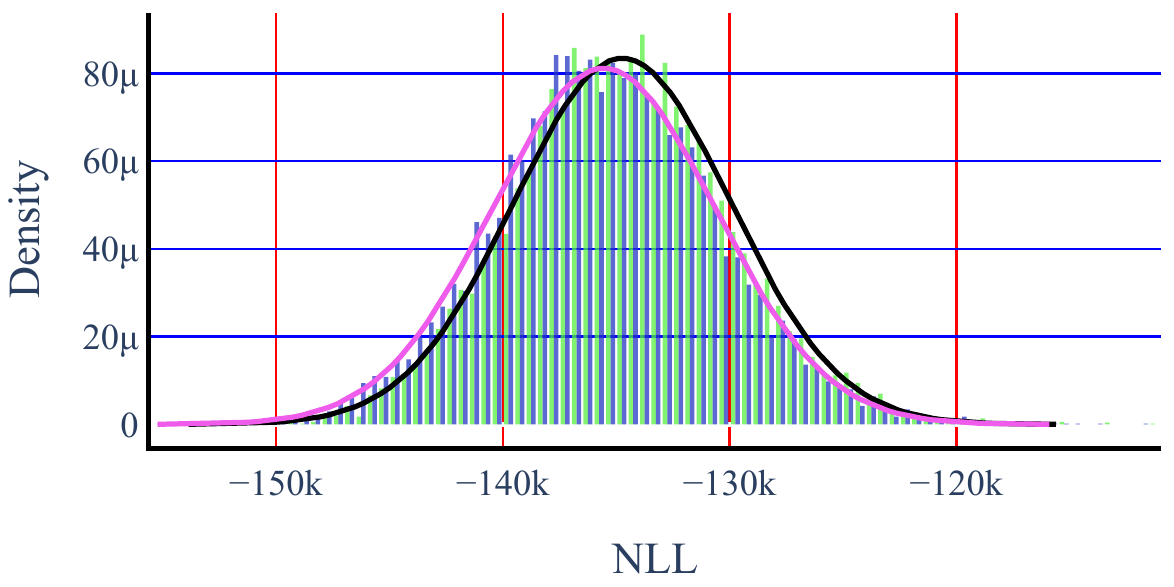}\tabularnewline
 \raisebox{23mm}{15}&\includegraphics[width=0.45\linewidth]{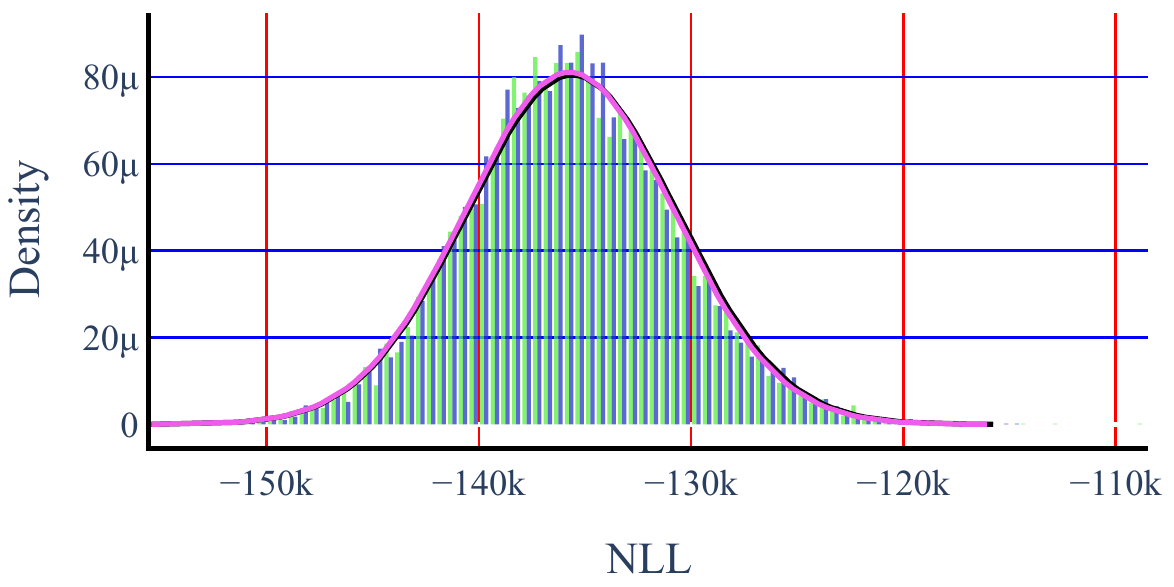} &
 \includegraphics[width=0.45\linewidth]{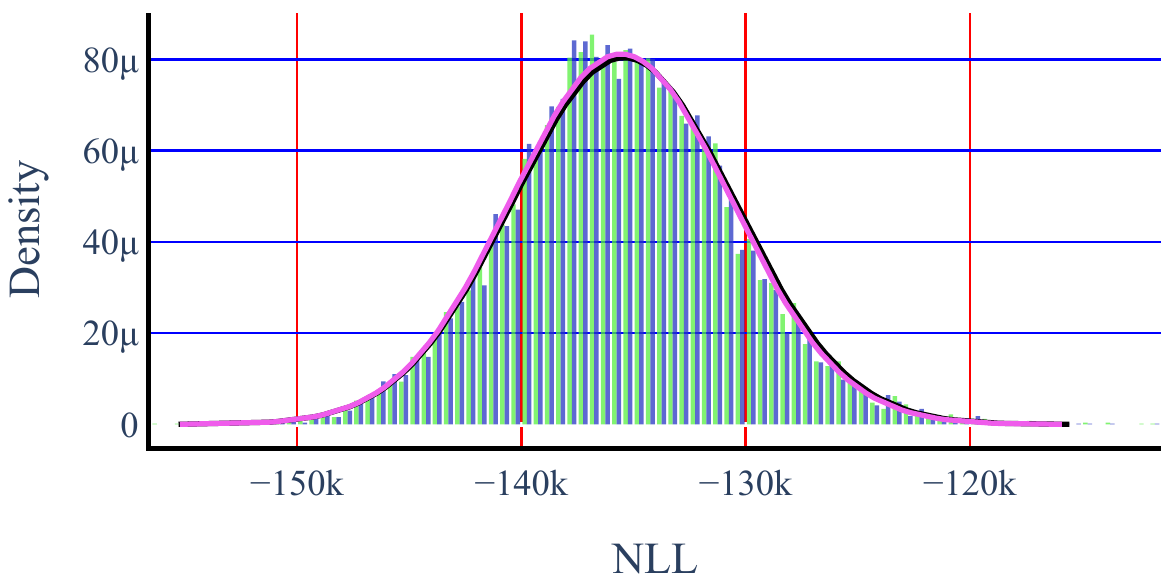}\tabularnewline
 &\multicolumn{2}{c}{\cbox{black} Tamed estimation ($\hat{\theta}_T$)\hspace{2mm} \cbox{dist_green} Tamed ($\theta_F$) \hspace{2mm} \cbox{dist_blue} Base ($\theta_B$) \hspace{2mm} \cbox{dist_pink} Base estimation ($\hat{\theta}_B$)}
\end{tabular}
\caption{\textbf{Preserving the NLL's distribution of the remember set $\D_R$.} Normalized density histogram and normal estimation of the NLL distribution on the base model training set, and a similar holdout set. The different plots correspond to models that were tamed to forget images of a specific identity, with a varying number of images. These plots suggest the distribution's change is minor.}
\label{supp:fig:preserve_dist}
\end{figure*}

\setcounter{figure}{0}
\setcounter{table}{0}
\section{Visualizations}\label{supp:sec:visualizations}
\begin{figure}

\let\ww\relax
\newlength{\ww}
\setlength{\ww}{0.23\linewidth}
\setlength{\tabcolsep}{1.2pt}
\renewcommand{\arraystretch}{1.0}
\centering
\begin{tabular}{@{}cc p{5pt} cc@{}}
\multicolumn{2}{c}{ Fairface}&\multicolumn{1}{c}{0}&\multicolumn{2}{c}{ CelebA validation}\tabularnewline
\includegraphics[frame,width=\ww,keepaspectratio]{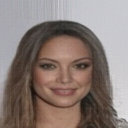}&
\includegraphics[frame,width=\ww,keepaspectratio]{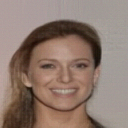}&

&

\includegraphics[frame,width=\ww,keepaspectratio]{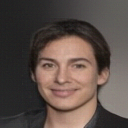}&
\includegraphics[frame,width=\ww,keepaspectratio]{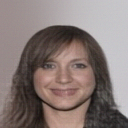}\tabularnewline

\includegraphics[frame,width=\ww,keepaspectratio]{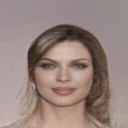}&
\includegraphics[frame,width=\ww,keepaspectratio]{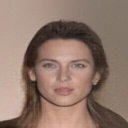}&
&

\includegraphics[frame,width=\ww,keepaspectratio]{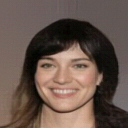}&
\includegraphics[frame,width=\ww,keepaspectratio]{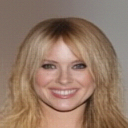}\tabularnewline
&&\multicolumn{1}{c}{20}\tabularnewline

\includegraphics[frame,width=\ww,keepaspectratio]{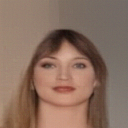}&
\includegraphics[frame,width=\ww,keepaspectratio]{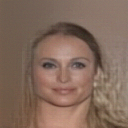}&
&

\includegraphics[frame,width=\ww,keepaspectratio]{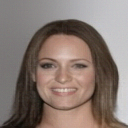}&
\includegraphics[frame,width=\ww,keepaspectratio]{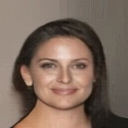}\tabularnewline
\includegraphics[frame,width=\ww,keepaspectratio]{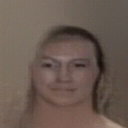}&
\includegraphics[frame,width=\ww,keepaspectratio]{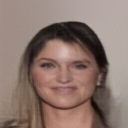}&
&
\includegraphics[frame,width=\ww,keepaspectratio]{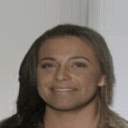}&
\includegraphics[frame,width=\ww,keepaspectratio]{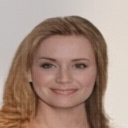}
\tabularnewline
&&\multicolumn{1}{c}{40}\tabularnewline
\includegraphics[frame,width=\ww,keepaspectratio]{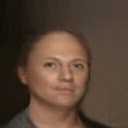}&
\includegraphics[frame,width=\ww,keepaspectratio]{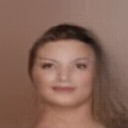}&
&

\includegraphics[frame,width=\ww,keepaspectratio]{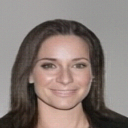}&
\includegraphics[frame,width=\ww,keepaspectratio]{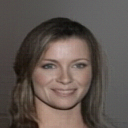}\tabularnewline

\includegraphics[frame,width=\ww,keepaspectratio]{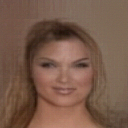}&
\includegraphics[frame,width=\ww,keepaspectratio]{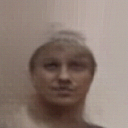}&
&
\includegraphics[frame,width=\ww,keepaspectratio]{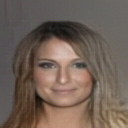}&
\includegraphics[frame,width=\ww,keepaspectratio]{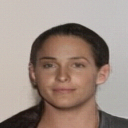}
\tabularnewline
\end{tabular}
\caption{\textbf{Taming with no training set.} Images generated using models that were tamed without any training data access, along the number of iterations of our method (0, 20, 40). The left side shows that as the remember set is more distant than the training set (Fairface~\cite{fairface}), the results are worse compared to unseen data from a closer distribution (CelebA validation set).}
\label{supp:fig:generate_no_data}
\end{figure}

In this section, we show different generated samples of tamed models from the different experiments in \cref{sec:experiments}.

We begin with \cref{supp:fig:generate_no_data}, showing the generated images when experimenting without any access to the training set as in \cref{sec:forget_without_train_set}.
This figure shows how the similarity between the remember set $\D_R$ and the training set $\D$ affects the generation quality, as when using a similar distribution (CelebA validation set) maintains generation quality while using a different distribution (FairFace~\cite{fairface}) does not.

\begin{figure*}

 \let\ww\relax
\newlength{\ww}
\setlength{\ww}{0.135\linewidth}

\setlength{\tabcolsep}{1.2pt}

 \let\wr\relax
\newlength{\wr}
\setlength{\wr}{0.1\linewidth}

\renewcommand{\arraystretch}{1.0}

\centering
\begin{tabular}{@{}cccccccc@{}}
        \multirow{3}{*}[-3em]{\raisebox{0.02\linewidth}{\rotatebox{90}{\small $-$ Blond}}}&\includegraphics[frame,width=\ww,keepaspectratio]{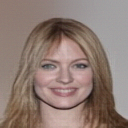}&\includegraphics[frame,width=\ww,keepaspectratio]{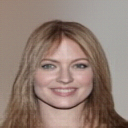}&\includegraphics[frame,width=\ww,keepaspectratio]{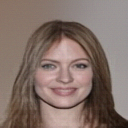}&\includegraphics[frame,width=\ww,keepaspectratio]{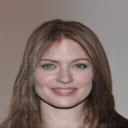}&\includegraphics[frame,width=\ww,keepaspectratio]{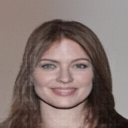}&\includegraphics[frame,width=\ww,keepaspectratio]{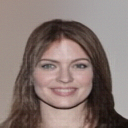}&\includegraphics[frame,width=\ww,keepaspectratio]{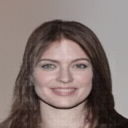}\tabularnewline
	&\includegraphics[frame,width=\ww,keepaspectratio]{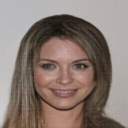}&\includegraphics[frame,width=\ww,keepaspectratio]{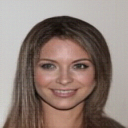}&\includegraphics[frame,width=\ww,keepaspectratio]{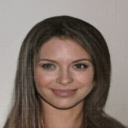}&\includegraphics[frame,width=\ww,keepaspectratio]{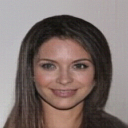}&\includegraphics[frame,width=\ww,keepaspectratio]{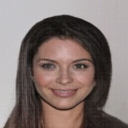}&\includegraphics[frame,width=\ww,keepaspectratio]{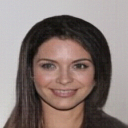}&\includegraphics[frame,width=\ww,keepaspectratio]{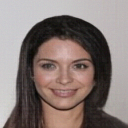}\tabularnewline
     &\includegraphics[frame,width=\ww,keepaspectratio]{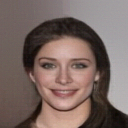}&\includegraphics[frame,width=\ww,keepaspectratio]{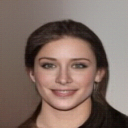}&\includegraphics[frame,width=\ww,keepaspectratio]{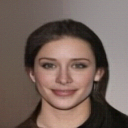}&\includegraphics[frame,width=\ww,keepaspectratio]{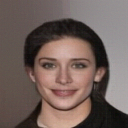}&\includegraphics[frame,width=\ww,keepaspectratio]{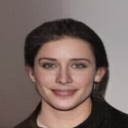}&\includegraphics[frame,width=\ww,keepaspectratio]{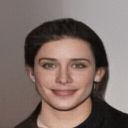}&\includegraphics[frame,width=\ww,keepaspectratio]{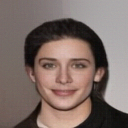}\tabularnewline
 
    \midrule

     \multirow{3}{*}[-2em]{\raisebox{0.02\linewidth}{\rotatebox{90}{\small $+$ Blond Smile}}}&\includegraphics[frame,width=\ww,keepaspectratio]{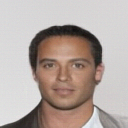}&\includegraphics[frame,width=\ww,keepaspectratio]{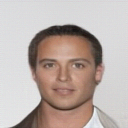}&\includegraphics[frame,width=\ww,keepaspectratio]{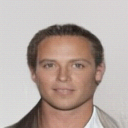}&\includegraphics[frame,width=\ww,keepaspectratio]{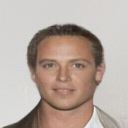}&\includegraphics[frame,width=\ww,keepaspectratio]{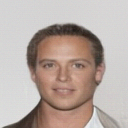}&\includegraphics[frame,width=\ww,keepaspectratio]{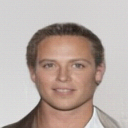}&\includegraphics[frame,width=\ww,keepaspectratio]{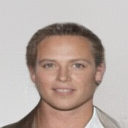}\tabularnewline
	&\includegraphics[frame,width=\ww,keepaspectratio]{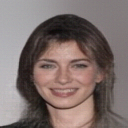}&\includegraphics[frame,width=\ww,keepaspectratio]{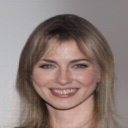}&\includegraphics[frame,width=\ww,keepaspectratio]{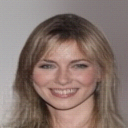}&\includegraphics[frame,width=\ww,keepaspectratio]{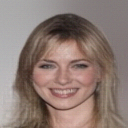}&\includegraphics[frame,width=\ww,keepaspectratio]{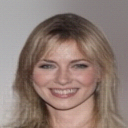}&\includegraphics[frame,width=\ww,keepaspectratio]{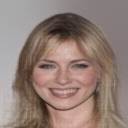}&\includegraphics[frame,width=\ww,keepaspectratio]{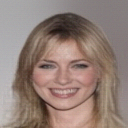}\tabularnewline
    &\includegraphics[frame,width=\ww,keepaspectratio]{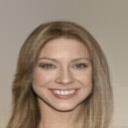}&\includegraphics[frame,width=\ww,keepaspectratio]{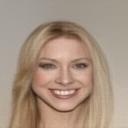}&\includegraphics[frame,width=\ww,keepaspectratio]{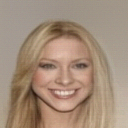}&\includegraphics[frame,width=\ww,keepaspectratio]{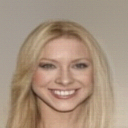}&\includegraphics[frame,width=\ww,keepaspectratio]{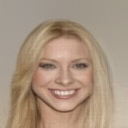}&\includegraphics[frame,width=\ww,keepaspectratio]{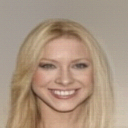}&\includegraphics[frame,width=\ww,keepaspectratio]{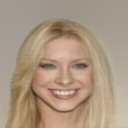}\tabularnewline
    
    \midrule

    \multirow{2}{*}{\raisebox{0.02\linewidth}{\rotatebox{90}{\small $+$ Male}}}&
    \includegraphics[frame,width=\ww,keepaspectratio]{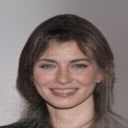}&\includegraphics[frame,width=\ww,keepaspectratio]{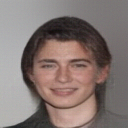}&\includegraphics[frame,width=\ww,keepaspectratio]{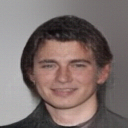}&\includegraphics[frame,width=\ww,keepaspectratio]{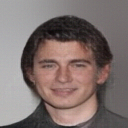}&\includegraphics[frame,width=\ww,keepaspectratio]{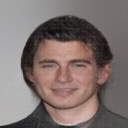}&\includegraphics[frame,width=\ww,keepaspectratio]{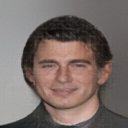}&\includegraphics[frame,width=\ww,keepaspectratio]{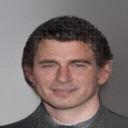}\tabularnewline

	&\includegraphics[frame,width=\ww,keepaspectratio]{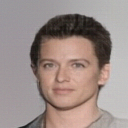}&\includegraphics[frame,width=\ww,keepaspectratio]{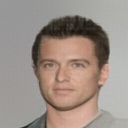}&\includegraphics[frame,width=\ww,keepaspectratio]{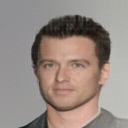}&\includegraphics[frame,width=\ww,keepaspectratio]{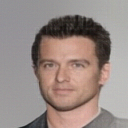}&\includegraphics[frame,width=\ww,keepaspectratio]{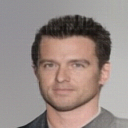}&\includegraphics[frame,width=\ww,keepaspectratio]{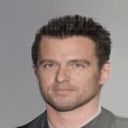}&\includegraphics[frame,width=\ww,keepaspectratio]{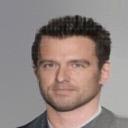}\tabularnewline
    
    &$0$&$25$&$50$&$75$&$100$&$125$&$150$
        
\end{tabular}
\caption{\textbf{Change attributes process.} By examining the same latent vectors during our process, we are able to visualize the change of different attribute (the bottom row includes the number of iterations).}
\label{supp:fig:change_attributes_combined}
\end{figure*}

\begin{figure*}

 \let\ww\relax
\newlength{\ww}
\setlength{\ww}{0.135\linewidth}

\setlength{\tabcolsep}{1.2pt}

 \let\wr\relax
\newlength{\wr}
\setlength{\wr}{0.01\linewidth}

\renewcommand{\arraystretch}{1.0}

\centering
\begin{tabular}{@{}cccccccc@{}}
       \raisebox{0.02\linewidth}{\rotatebox{90}{\small $+$ Beard}}&\includegraphics[frame,width=\ww,keepaspectratio]{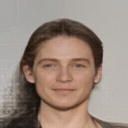}&\includegraphics[frame,width=\ww,keepaspectratio]{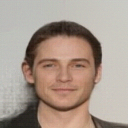}&\includegraphics[frame,width=\ww,keepaspectratio]{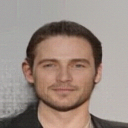}&\includegraphics[frame,width=\ww,keepaspectratio]{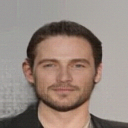}&\includegraphics[frame,width=\ww,keepaspectratio]{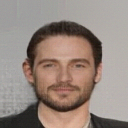}&\includegraphics[frame,width=\ww,keepaspectratio]{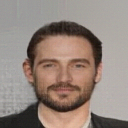}&\includegraphics[frame,width=\ww,keepaspectratio]{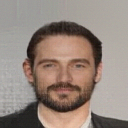}\tabularnewline
        \raisebox{0.005\linewidth}{\rotatebox{90}{\small $-$ Open Mouth}}&\includegraphics[frame,width=\ww,keepaspectratio]{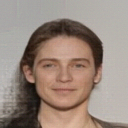}&\includegraphics[frame,width=\ww,keepaspectratio]{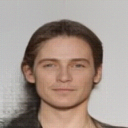}&\includegraphics[frame,width=\ww,keepaspectratio]{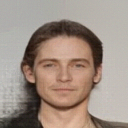}&\includegraphics[frame,width=\ww,keepaspectratio]{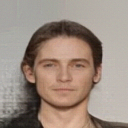}&\includegraphics[frame,width=\ww,keepaspectratio]{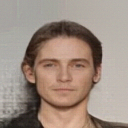}&\includegraphics[frame,width=\ww,keepaspectratio]{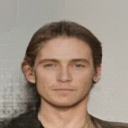}&\includegraphics[frame,width=\ww,keepaspectratio]{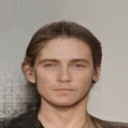}\tabularnewline
        \raisebox{0.01\linewidth}{\rotatebox{90}{\small $+$ Eyeglasses}}&\includegraphics[frame,width=\ww,keepaspectratio]{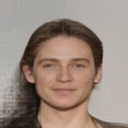}&\includegraphics[frame,width=\ww,keepaspectratio]{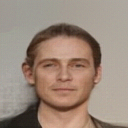}&\includegraphics[frame,width=\ww,keepaspectratio]{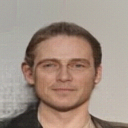}&\includegraphics[frame,width=\ww,keepaspectratio]{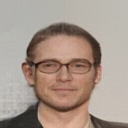}&\includegraphics[frame,width=\ww,keepaspectratio]{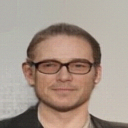}&\includegraphics[frame,width=\ww,keepaspectratio]{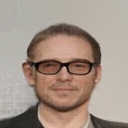}&\includegraphics[frame,width=\ww,keepaspectratio]{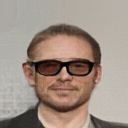}\tabularnewline
        \raisebox{0.03\linewidth}{\rotatebox{90}{\small $+$ Bald}}&\includegraphics[frame,width=\ww,keepaspectratio]{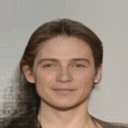}&\includegraphics[frame,width=\ww,keepaspectratio]{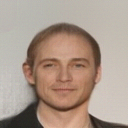}&\includegraphics[frame,width=\ww,keepaspectratio]{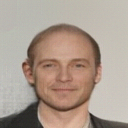}&\includegraphics[frame,width=\ww,keepaspectratio]{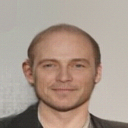}&\includegraphics[frame,width=\ww,keepaspectratio]{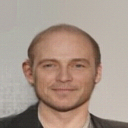}&\includegraphics[frame,width=\ww,keepaspectratio]{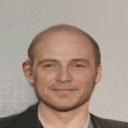}&\includegraphics[frame,width=\ww,keepaspectratio]{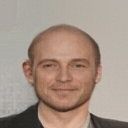}\tabularnewline
        &$0$&$25$&$50$&$75$&$100$&$125$&$150$
\end{tabular}
\caption{\textbf{Change different attributes of a single identity.} We visualize the change of a single identity by visualizing the changes in \emph{different} experiments on the \emph{same} latent vector (the bottom row includes the number of iterations).}
\label{supp:fig:same_identity_change_attrbiutes}
\end{figure*}

\cref{supp:fig:change_attributes_combined,supp:fig:same_identity_change_attrbiutes} Show additional examples of taming an attribute (\cref{sec:forget_attribute}), as demonstrated in \cref{supp:fig:same_identity_change_attrbiutes}. 

In \cref{supp:fig:change_attributes_combined}, we see that an identity possessing blond hair can quickly be scrubbed of that attribute ($1^{\textrm{st}}$ row).
Identities without blond hair will obtain a darker hair color as a result of this process, as we globally reduce the blond hair attribute ($2^{\textrm{nd}}$ row).
This property can be used to debias a model, \eg, a model that generates images of females with higher probability, can be tamed in order to achieve a higher generation probability of males (as shown in the $7^{\textrm{th}}$ row). 
This figure also shows that the changes are related to the data in the forget and remember sets. This can be seen in the $4^{\textrm{th}}$ row, as the blond hair change on the male identity is less impactful compared to the female ones. This is due to the fact the training data (CelebA) only has $0.85\%$ images of blond males.

\cref{supp:fig:same_identity_change_attrbiutes} shows how while we change an attribute globally, when we focus on a single latent vector, even in different experiments, the attribute change is applied while preserving the original identity. 


\end{document}